\documentclass[runningheads]{llncs}

\usepackage{eccv}

\usepackage{eccvabbrv}

\usepackage{graphicx}
\usepackage{booktabs}
\usepackage{array}
\usepackage{caption}
\usepackage{tabularx}
\usepackage{makecell}
\usepackage{multirow}
\usepackage{amsmath}
\usepackage{amsfonts}
\usepackage{tikz}
\usepackage{subcaption}
\usepackage{dashrule}
\usepackage{caption}
\usepackage{placeins}
\usepackage{ulem}
\usepackage[accsupp]{axessibility}  
\newcommand{\yjc}[1]{\textcolor{black}{{#1}}}
\newcommand{\lqf}[1]{\textcolor{black}{{#1}}}

\newcommand{\benchmarkname}{\textbf{CornerCaseRepo}}
\newcommand{\rulesep}{\unskip\ \vrule\ }

\usepackage{hyperref}

\usepackage{orcidlink}

\begin{document}

\title{Think2Drive: Efficient Reinforcement Learning by Thinking with Latent World Model for Autonomous Driving (in CARLA-v2)} 

\titlerunning{Think2Drive}


\author{Qifeng Li\inst{*}\orcidlink{0000-0003-3813-0778} \and
Xiaosong Jia\inst{*}\orcidlink{0000-0002-5222-1476} \and
Shaobo Wang \and 
Junchi Yan\inst{\dagger}\orcidlink{0000-0001-9639-7679}}

\authorrunning{Q.~Li, X.~Jia et al.}

\institute{Shanghai Jiao Tong University, Shanghai 200240, China \\
\email{\{liqifeng, jiaxiaosong, shaobowang1009, yanjunchi\}@sjtu.edu.cn} \\
$^*$ Equal contributions \quad
$^\dagger$Correspondence (partly supported by NSFC 92370201)\\
}

\maketitle
\begin{abstract}
  Real-world autonomous driving (AD) like urban driving involves many corner cases. The lately released AD Benchmark CARLA Leaderboard v2 (a.k.a. CARLA v2) involves 39 new common events in the driving scene, providing a more quasi-realistic testbed compared to CARLA Leaderboard v1. It poses new challenges and so far no literature has reported any success on the new scenarios in V2. In this work, we take the initiative of directly training a neural planner and the hope is to handle the corner cases flexibly and effectively. To our best knowledge, we develop the first model-based RL method (named Think2Drive) for AD, with a compact latent world model to learn the transitions of the environment, and then it acts as a neural simulator to train the agent i.e. planner. It significantly boosts the training efficiency of RL thanks to the low dimensional state space and parallel computing of tensors in the latent world model.  Think2Drive is able to run in an expert-level proficiency in CARLA v2 within 3 days of training on a single A6000 GPU, and to our best knowledge, so far there is no reported success (100\% route completion) on CARLA v2. We also develop CornerCaseRepo, a benchmark that supports the evaluation of driving models by scenarios. We also propose a balanced metric to evaluate the performance by route completion, infraction number, and scenario density.
  \keywords{Autonomous driving \and Neural planner \and World model \and Model-based reinforcement learning \and CARLA v2 \and Think2Drive}
\end{abstract}


\textit{\small{We are getting to the point where there’s one last piece of the system that needs to be a neural net which is the planning and control function. }}

\rightline{\textit{\small{Elon Musk, 2023 Tesla Annual Shareholder Meeting}}}

\textit{\small{Vehicle control is the final piece of the Tesla FSD AI puzzle. That will drop >300k lines of C$++$ control code by ~2 orders of magnitude. It is training as I write this.}} 
 
\rightline{\small{\textit{Elon Musk, Twitter August, 2023}}}


\section{Introduction}
\label{sec:intro}
Autonomous driving (AD)~\cite{li2022bevsurvey,hu2023uniad,yang2023survey}, especially urban driving, requires the vehicles to engage with dense and diverse traffic participants~\cite{raljia,pmlr-v205-jia23a,Jia2022HDGTHD} and adapt to complex and dynamic traffic scenarios. Traditional manually-crafted rule-based planning methods struggle to handle these scenarios due to their reliance on exhaustive coverage of all cases, which is nearly impossible for long-tail scenarios~\cite{karnchanachari2024towards,lu2024activead}. Additionally, ensuring compatibility between new and existing rules becomes increasingly challenging as the decision tree expands~\cite{man-month}. As a result, \textbf{there is a trend that to adopt neural planner, offering hope for scaling up with data and computation to achieve full driving autonomy.}


There have emerged benchmarks for the development and validation of planning methods in AD, e.g. HighwayEnv~\cite{highway-env} and CARLA Leaderboard v1 (a.k.a. CARLA v1)~\cite{dosovitskiy2017CARLA}.  However, in these pioneering benchmarks, the behaviors of environment agents are usually simple and the diversity and complexity of road conditions are often limited, bearing a gap to real-world driving. In fact, most of their tasks can be effectively addressed by rule-based approaches~\cite{Chitta2023PAMI} via basic skills like lane following, adherence to traffic signs, and collision avoidance, which is however well below the difficulty level of real-world urban driving. For instance, in CARLA v1, the rule-based Autopilot with only hundreds of lines of code could achieve nearly perfect performance~\cite{Jaeger2023ICCV}. Thus, methods developed for or verified in these environments are possibly unable to handle many common real-world traffic scenarios, which significantly limits their practical value.

A quasi-realistic benchmark, CARLA Leaderboard v2 (a.k.a. CARLA v2)~\cite{CARLAv2} was released in November 2022, encompassing 39 real-world corner cases in addition to its v1 version with 10 cases. For instance, there are scenarios where the ego vehicle is on a two-way single-lane road and encounters a construction zone ahead. It requires the ego agent to invade the opposite lane when it is sufficiently clear, circumventing the construction area, and promptly merging back into the original lane afterward. In particular, corner cases, as their names suggest, are sparse in both the real world and the routes provided by CARLA v2, posing a long-tail problem for learning. 

Being aware of the difficulty of the new benchmark, the CARLA team also provides several human demonstrations of completing these scenarios. Though humans could effortlessly navigate such scenarios, it is highly non-trivial to write into rules, not to mention adopting the popular imitation learning methods~\cite{hussein2017imitation} with few samples, as succeed in CARLA v1. Due to the much-increased difficulty, widely used rule-based experts like autopilot~\cite{Jaeger2023ICCV} and learning-based experts like Roach~\cite{zhang2021roach} (by model-free RL) both can not work in CARLA v2 at all. Up to date, there is no success reported on CARLA v2, one year after its release.


In this paper, we aim to obtain the driving policy under such a quasi-realistic AD benchmark by learning, ambitiously with a model-based RL~\cite{pmlr-v97-hafner19a} approach which hopefully would enjoy two merits: data efficiency and flexibility against complex scenarios. Note that developing model-based RL can be nontrivial, w.r.t. the specific domain. It not only involves difficulties intrinsic to AD such as complex road conditions and highly interactive behaviors in long-tailed distributions, but also engineering problems of CARLA e.g. collecting massive samples efficiently from a cumbersome simulator. In contrast, to our best knowledge, existing RL methods for AD~\cite{chekroun2023gri,zhang2021roach} are mostly model-free~\cite{schulman2017proximal,mnih2013playing}, which would inherently suffer from data inefficiency in CARLA v2 due to its complexity. 
There are also a few model-based RL methods applied to AD~\cite{henaff2019model,diehl2021umbrella} in simple benchmarks, but far from solving CARLA v2.

Specifically, we model the environment's transition function in AD using a world model~\cite{ha2018world,hafner2019learning,hafner2023mastering} and employ it as a neural network simulator, to make the planner `think' to drive (\textbf{Think2Drive}) in the learned latent space. 

In this way, the data efficiency could be significantly increased since the neural network could in parallel conduct hundreds of rollouts with a much faster iteration speed compared to the physical simulator i.e. CARLA. However, even with the state-of-the-art model-based methods~\cite{hafner2023mastering}, it is still highly non-trivial to adopt them for AD which has its unique characteristics as mentioned below compared to Atari or MineCraft as done in \cite{hafner2023mastering}.
Specifically, we consider three major obstacles in model-based RL for quasi-realistic AD.
 



\textbf{1) Policy degradation.} There might exist contradictions among optimal policies of different scenarios. For instance, for scenario I where the front vehicle suddenly brakes, and scenario II where the planner has to merge into high-speed traffic, the former requires the planner to keep a safe distance from the preceding vehicle, while the latter demands proactive engagement with the front vehicles. 
Consequently, the driving model will be easily trapped in the local optima. 
To mitigate this issue, we randomly re-initialize all weights of the planner in the middle of training while keeping the world model unchanged inspired by~\cite{nikishin2022primacy}, allowing the planner to escape local optima for policy degradation preventing. As the world model can provide the planner with accurate and dense rewards, the reinitialized planner can better deal with the cold-start problem.

\textbf{2) Long-tail nature.}  
As mentioned above, the long-tail nature of AD tasks poses a significant challenge for the planner to handle all the corner cases. We implement an automated scenario generator that can generate scenarios based on road situations, thus providing the planner with abundant, scenario-dense data. 
We further design a termination-priority replay strategy, ensuring that the world model and planner prioritize exploration on long-tailed valuable states. 

\textbf{3) Vehicle heading stabilization.} For a learning-based planner, maintaining the same action over a long time is hard. However, stability and smoothness of control are required in the context of autonomous driving, such as maintaining a steady steer value on a straight lane. Therefore, we also introduce a steering cost function to stabilize the vehicle's heading. 

Beyond these three major obstacles, the training of a model-based AD planner also encounters challenges such as initial running difficulties, delayed learning signals, etc. We address them brick by brick, with \textbf{a detailed discussion provided in~\cref{subsec:bricks}}. 
By developing all the above techniques, we manage to establish our model, Think2Drive, which has achieved \textbf{the pioneering feat of successfully addressing all 39 quasi-realistic scenarios within 3 days of training on a single GPU A6000}. Think2Drive can also serve as a planning module or teacher model for learning-based driving models.

\textbf{The highlights of the paper are as follows.} 
1) To our best knowledge, it is the first model-based RL approach for AD (i.e. neural planner) in literature that manages to handle quasi-realistic scenarios with techniques like \textbf{resetting technique, automated scenario generation, termination-priority replay strategy, steering cost function}, etc. 
2) We propose a new and balanced metric to evaluate the performance by route completion, infraction number and scenario density. 
3) Experimental results on CARLA V2 and the proposed \benchmarkname\ benchmark show the superiority of our approach. 
A demo of our proficient planner on CARLA V2 test routes is available at 
\url{https://thinklab-sjtu.github.io/CornerCaseRepo/}


\section{Related Works}
\textbf{Model-based Reinforcement Learning}.
Model-based reinforcement learning explicitly utilizes a world model to learn the transition of the environment and make the actor purely interact with the world model to improve data efficiency. PlaNet~\cite{hafner2019learning} proposes the recurrent state-space model (RSSM) to model both the deterministic and stochastic part of the environment followed by many later works~\cite{hafner2019dream, hafner2020mastering, schrittwieser2020mastering,hafner2023mastering}. For instance, Dreamer~\cite{hafner2019dream}, Dreamer2~\cite{hafner2020mastering}, and Dreamer3~\cite{hafner2023mastering} progressively improve the performance of the world model based on RSSM. Notably, DreamerV3 achieves state-of-the-art performance across multiple tasks including Minecraft and Crafter, without the need of parameter tuning by employing techniques including symlog loss and free bits. 
Daydreamer~\cite{wu2023daydreamer} further extends the application of model-based approaches to physical robots. 

We note that model-based RL is especially fit for AD since 1) the super data efficiency of the model-based method could be the key to deal with the long-tailed issue of AD while the physical simulator is usually burdensome; 2) the transition of AD scenes under rasterized BEV is relatively easy to learn compared to Atari or MineCraft, which means one would be able to train an accurate world model. 

\textbf{Reinforcement Learning-based Agents in CARLA}.
Reinforcement learning is an important technique to obtain planning agents in CARLA, which could serve as expert models. \cite{Chen_Yuan_Tomizuka_2019} explores the utilization of BEV data as input for DDQN \cite{Van_Hasselt_Guez_Silver_2022}, TD3 \cite{Fujimoto_Hoof_Meger_2018}, and SAC \cite{Haarnoja_Zhou_Abbeel_Levine_2018}, with the added step of pre-training the image encoder on expert trajectories. \cite{Rhinehart_McAllister_Levine_2018} investigates the integration of IL with reinforcement learning. MaRLn~\cite{toromanoff2020end} uses raw sensor inputs to train a reinforcement learning (RL) agent, failing to achieve good performance in Leaderboard v1.  Roach~\cite{zhang2021roach} is the state-of-the-art model-free RL agent widely used as the expert model of recent end-to-end AD~\cite{wu2022trajectory,wu2023PPGeo,jia2023think,jia2023driveadapter}, yet it fails to handle the CARLA v2 (details in~\cref{subsec:performance}). 
There are also a few model-based RL methods applied to AD~\cite{henaff2019model,diehl2021umbrella} in simple benchmarks, but the difficulty is far from CARLA v2. Notably, after the publishing of Think2Drive, PDM-lite~\cite{pdm-lite}, a recently released rule-based planner (after our arxiv version \cite{li2024think2driveefficientreinforcementlearning}), is able to solve all scenarios in CARLA Leaderboard v2 as well. However, they adopt different hyper-parameters for different scenarios, which requires heavy manual labor.

\section{Methodology}
\subsection{Problem Formulation with Model-based RL}
As our focus is on planning, we use the input of privileged information $x_t$ including bounding boxes of surrounding agents and obstacles, HD-Map, states of traffic lights, etc, eliminating the influence of perception. The required output is the control signals: $A: \text{throttle, steer, brake}$. 

We construct a planner model $\pi_\eta$ which outputs action $a_t$ based on current state $s_t$ and construct a world model $F$ to learn the transition of the driving scene so that the planner model could drive and be trained by ``think" instead of directly interacting with the physical simulator. The iteration process of ``think" is as follows: given an initial input $x_t$ at time-step $t$ sampled from the record, the world model encodes it as state $s_t$. Then, the planner generates $a_t$ based on $s_t$. Finally, the world model predicts the reward $r_{t}$, termination status $c_{t}$, and the future state $s_{t+1}$ with $s_t$ and $a_t$ as input. The overall pipeline is: 
\begin{equation}
    \label{eq:basic_model}
    \begin{aligned}
        s_t &\leftarrow F_\theta^{Enc}(x_t), \quad
        a_t &\leftarrow \pi(s_t), \quad
        s_{t+1} &\leftarrow F_\theta^{Pre}\left(s_t, a_t\right) 
    \end{aligned}
\end{equation}
By rollouting in the latent state space $s_t$ of the world model, the planner can think and learn efficiently, without interacting with the heavy physical simulator. 
\subsection{World Model Learning and Planner Learning}
We use DreamerV3~\cite{hafner2023mastering}'s structure and objective to train the world model and planner model. \yjc{Note that our main novelty lies in the first successful adoption of latent world model to AD.}


\textbf{World Model Learning}. It has four components in line with~\cite{hafner2023mastering}: 
\begin{equation}
\hspace{-0.1cm}
\begin{aligned}
    \operatorname{RSSM} 
    & \begin{cases}
        \text { Sequence model: } & \quad h_t=f_\theta\left(h_{t-1}, z_{t-1}, a_{t-1}\right) \\ 
        \text { Encoder: } & \quad z_t \sim q_\theta\left(z_t \mid h_t, x_t\right) \\ 
        \text { Dynamics predictor: } & \quad \hat{z}_t \sim p_\theta\left(\hat{z}_t \mid h_t\right) \\ 
    \end{cases} \\
    & \hspace{0.4cm} \begin{aligned}
        & \text { Reward predictor: } && \hat{r}_t \sim p_\theta\left(\hat{r}_t \mid h_t, z_t\right)  \\ 
        & \text { Termination predictor: } && \hat{c}_t \sim p_\theta\left(\hat{c}_t \mid h_t, z_t\right) \\ 
        & \text { Decoder: } && \hat{x}_t \sim p_\theta\left(\hat{x}_t \mid h_t, z_t\right)\\
    \end{aligned}
\end{aligned}
\end{equation}
where \textbf{RSSM is to provide an accurate transition function of the environment in latent space and perform efficient rollouts for the planner model}. 
It decomposes the state representation $s_t$ into stochastic representation $z_t$ and deterministic hidden state $h_t$ based on~\cref{eq:basic_model} to better model the corresponding deterministic and stochastic aspects of the true transition function. 
The encoder first maps raw input $x_t$ to latent representation $z_t$, then the sequence model predicts the future hidden state $h_{t+1}$ based on the representation $z_t$, action $a_t$, and history hidden state $h_{t}$. 
\textbf{The reward predictor forecasts the reward  $r_{t}$ associated with the model state $s_{t}=(h_{t}, z_{t})$ and the termination predictor predicts the termination flags $c_t \in \{0,1\}$, which both provide learning signals for the planner model.} 
The decoder reconstructs inputs to ensure informative representation and generate interpretable images. Specifically, the world model's training loss~\cite{hafner2023mastering} consists of:
\begin{equation}
     \begin{aligned}
        \mathcal{L}_{\text{pred}}(\theta) \doteq&-\ln p_\theta\left(x_t \mid z_t, h_t\right)-\ln p_\theta\left(r_t \mid z_t, h_t\right)-\ln p_\theta\left(c_t \mid z_t, h_t\right) \\
        \mathcal{L}_{\text{dyn}}(\theta)  \doteq& \max \left(1, \operatorname{KL}\left[\operatorname{fz}\left(q_\theta\left(z_t \mid h_t, x_t\right)\right) \| p_\theta\left(z_t \mid h_t\right)\right]\right) \\
        \mathcal{L}_{\text{rep}}(\theta)  \doteq& \max \left(1, \operatorname{KL}\left[q_\theta\left(z_t \mid h_t, x_t\right) \| \operatorname{fz}\left(p_\theta\left(z_t \mid h_t\right)\right)\right]\right)
    \end{aligned} 
\end{equation}
where the prediction loss $\mathcal{L}_{\text{pred}}$ trains both the decoder and the termination predictor via binary cross-entropy. 
Symlog loss~\cite{hafner2023mastering} is utilized to train the reward predictor. 
By minimizing the KL divergence between the prior  $p_\theta\left(z_t \mid h_t\right)$, the dynamics loss $\mathcal{L}_{\text{dyn}}$ trains the sequence model to predict the next representation, and the representation loss $\mathcal{L}_{\text{rep}}$ is used to lower the difficulty of this prediction. 
The two losses differ in the position of parameter-freeze operation $\operatorname{fz}(\cdot)$. 

Given a rollout of $x_{1:T}$, actions $a_{1:T}$, rewards $r_{1:T}$, and termination flag $c_{1:T}$ from records, the overall loss is:
\begin{equation}
  \label{equ:world-model-loss}
        \mathcal{L}(\theta) \doteq \mathrm{E}_{q_\theta}  \sum_{t=1}^T\left(\beta_{\mathrm{pred}} \mathcal{L}_{\mathrm{pred}}^t(\theta)+\beta_{\mathrm{dyn}} \mathcal{L}_{\mathrm{dyn}}^t(\theta)  
         + 
    \beta_{\mathrm{rep}} \mathcal{L}_{\mathrm{rep}}^t(\theta)\right)
\end{equation}

\textbf{Planner Learning.} 
The planner is learned via an actor-critic~\cite{sutton1999policy} architecture, where the planner model serves as the actor and a critic model is constructed to assist its learning. Benefiting from the world model, the planner model can purely think to drive in the latent space with high efficiency.  Specifically, given an input $x_t$ at $t$ from the record as a start point, the world model first maps it to $s_t=\left(z_t, h_t\right)$. Then, the world model and the planner model conduct $T$ steps exploration: $\left<\hat{s}_{1:T}, a_{0:T}, r_{0:T}, c_{0:T}\right>$. 
The planner model $\pi_\eta\left(a|s\right)$ tries to maximize the expected discounted return generated by the reward predictor: $\sum_t \gamma \hat{r}_t$ while the critic learns to evaluate each state conditioned on the planner's policy: $V(s_T) \approx \mathrm{E}_{s \sim F_\theta, a \sim \pi_\eta}\left(R_t\right)$. 

To handle the accumulated error over the horizon $T$, the expected return is clipped by $T=15$ and the left return is estimated by the critic: $R_{T:} = v(s_T)$. We follow DreamerV3 by employing the bucket-sorting rewards and two-hot encoding to stably train the critic. Given the two-hot encoded target $y_t = \operatorname{fz}(\text{twohot}(\text{symlog}(R_t^\lambda))))$, the cross-entropy loss is used to train the critic~\cite{hafner2023mastering}: 
\begin{equation}
    \mathcal{L}_{\text {critic }}(\psi) \doteq -\sum_{t=1}^T y_t^\top \ln p_\psi\left(\cdot \mid s_t\right)
\end{equation}
where the softmax distribution $p_\psi\left(\cdot \mid s_t\right)$ over equal split buckets is the output of the critic. 
The reward expectation term for actor training is normalized by moving statistics~\cite{hafner2020mastering,williams1992simple}: 
\begin{equation}
    \mathcal{L}(\theta) \doteq \sum_{t=1}^T \left( \mathrm{E}_{\pi_\eta, p_\theta}\left[\frac{\operatorname{fz}\left(R_t^\lambda\right)}{\max (1, S)}\right]
    -\beta_{en} \mathrm{H}\left[\pi_\eta\left(a_t \mid s_t\right)\right] \right)
\end{equation}
where $S$ is the decaying mean of the range from their $5^{th}$ to the $95^{th}$ batch percentile. More details can be found in~\cite{hafner2023mastering}.

\subsection{Challenges and Our Devised Bricks}
\label{subsec:bricks}
After having the training paradigm determined, it is still not ready to solve the problem, as the autonomous driving scene has very different characteristics compared to Atari or MineCraft, e.g. policy degradation(\textit{Challenge 1}), long-tail nature(\textit{Challenge 2,3}), car heading stabilization(\textit{Challenge 4}), and some other obstacles(\textit{Challenge 5,6,7}). 
These obstacles make it highly non-trivial to adopt MBRL for AD. We devise essential Bricks to address them one by one. 

\textit{Challenge 1}: 
During training, the agent may be trapped in the local optimal policy of easy scenarios. 
This issue arises from potential contradictions in optimal strategies required for different scenarios. 
For example, in scenario I where the front vehicle suddenly brakes and scenario II where the planner has to merge into high-speed traffic, the former requires the planner to keep a safe distance from the preceding vehicle, while the latter demands proactive engagement with the front vehicles.  
Since the former is much easier than the latter, the model can be easily trapped into the local optima of keeping safe distance. 



\textbf{Brick 1}: 
We leverage the reset technique~\cite{nikishin2022primacy} in the middle of the training process where 
we randomly re-initialize all the parameters of the planner, allowing it to escape from the local optima. Notably, different from those model-free methods, the cold-start problem of the reset trick is less damaging since we have the well-trained world model to provide dense rewards.


\textit{Challenge 2}: 
The 39 scenarios are sparse in the released routes of CARLA v2. It brings a long-tail problem, incurring skewed exploration over trivial states. Also, the released scenarios are coupled with specified waypoints. As a result, the scenarios happen in a few fixed locations with limited diversity. 


\textbf{Brick 2}:
We implement an automated scenario generator.  Given a route, it can automatically split the route into multiple short routes and generate scenarios according to the road situation. 
As a result, the training process is able to acquire numerous shorter routes with dense scenarios. Besides, we build a benchmark \benchmarkname\ for evaluation which could estimate the detailed capabilities under each scenario while the official test routes are too long and thus their results are difficult to analyze. 

\textit{Challenge 3}: 
Valuable transitions occur non-uniformly over time and thus it is inefficient to train the world model under uniform sampling.
For example, in a 10-second red traffic light, 
an optimal policy with a decision frequency of 10 FPS will produce 100 consecutive frames of low-value transition, indicating that the exploration space of the planner remains highly imbalanced and long-tailed. 

\textbf{Brick 3}:
One kind of valuable transition can be easily located, i.e. the $K$ frames preceding the termination frame of an episode. 
Such terminations are either due to biases in the world model or exploration behavior, 
both of which could be especially valuable for the world model to learn the transition functions. 
Consequently, we employ a termination-priority sampling strategy where we either randomly sample or sample at the termination state with equal probability.

\textit{Challenge 4}:  
For reinforcement learning agents, particularly stochastic agents, maintaining a consistent action over an extended trajectory presents a challenge. For example, as could be observed in the demo of Roach~\cite{zhang2021roach}, the head of the ego vehicle would fluctuates even when driving in the straight road.  However, this consistency is often a requisite in the context of autonomous driving. 

\textbf{Brick 4}: 
We incorporate a steering cost function into the training of our agent. 
This cost function has enabled our model to achieve stable navigation.


\textit{Challenge 5}: 
The difficulty varies across scenarios. Directly training an agent with all scenarios may result in an excessively steep learning curve. Specifically, for these safety-critical scenarios requiring subtle control of the ego vehicles, it has a high risk of having violations or collisions and thus the model would be trapped in the over-conservation local optima (as evidenced in \cref{subsec:ablation}). 


\textbf{Brick 5}: 
Inspired by curriculum learning~\cite{bengio2009curriculum}, prior to undertaking unified training across all scenarios, we conduct a warm-up training stage for the RL model, using simple lane following and simple-turn scenarios so that the model has the basic driving skills and then we let it deal with these complex scenarios.

\textit{Challenge 6}: 
The dynamics of the driving environment are relatively stable compared to many stochastic tasks. 
The world model can gradually learn a more accurate transition function of the environment and generate more precise rewards for the planner. 
While the agent network relies on delayed, world model-generated rewards, requiring a longer time to converge. 
If the training ratio for both the world model and the agent is set equivalently, as is common in many tasks, this could decelerate the training process. 

\textbf{Brick 6}:
We set an incremental train ratio for the planner model where at the end the train ratio of the planner will be four times that of the world model, to expedite the convergence speed. 

\textit{Challenge 7}: 
RL training necessitates efficient exploration and environment resets, 
while the time cost of every reset in CARLA is unacceptable ($>40 s$ to load the route and instantiate all scenarios). 

\textbf{Brick 7}:
We wrap CARLA as an RL environment with standardized APIs and boost its running efficiency via asynchronous reloading and parallel execution. 
Details can be found in~\cref{fig:benchmark}, \cref{subsec:ablation}

\begin{table}[tb!]
  \centering
  \caption{\textbf{Driving performance and infraction of agents on the proposed \benchmarkname\ benchmark}. Mean and standard deviation are over 3 runs.}
  \resizebox{1.0\textwidth}{!}{
  \begin{tabular}{@{}c|c|ccccccccccc@{}}
    \toprule
    \makecell{Input} & Method & \makecell{Driving\\score} & \makecell{Weighted\\DS} & \makecell{Route\\completion} & \makecell{Infraction\\penalty} & \makecell{Collision\\pedestrians} & \makecell{Collision\\vehicles} & \makecell{Collision\\layout} & \makecell{Red light\\Infraction} & \makecell{Stop sign\\infraction} & \makecell{Agent\\blocked}\\
    \midrule
    \multirow{2}{*}[-4pt]{\makecell{Privileged\\Information}}&Roach~\cite{zhang2021roach} & 57.5$\pm$9 & 54.8$\pm$0.5 & 96.4$\pm$1.1 & 0.59$\pm$0.28 & 0.85$\pm$0.56 & 8.42$\pm$4.65 & 0.85$\pm$0.51 & 0.56$\pm$0.45 & 0.49$\pm$0.44 & 0.78$\pm$0.31\\
    &\textbf{\makecell{Think2Drive\\(Ours)}} & \textbf{83.8$\pm$1} & \textbf{89.0$\pm$0.2}  & \textbf{99.6$\pm$0.1} & \textbf{0.84 $\pm$ \textbf{0.01}} & \textbf{0.16$\pm$0.01} & \textbf{1.2$\pm$0.5} & \textbf{0.29$\pm$0.02} & \textbf{0.14$\pm$0.01} & \textbf{0.03$\pm$0.01} & \textbf{0.08$\pm$0.01} \\ \midrule\midrule
    Raw Sensors & \makecell{Think2Drive\\+TCP~\cite{wu2022trajectory}} & 36.40$\pm$12.23  & 29.6$\pm$0.2 & 85.88$\pm$8.26 &  0.41$\pm$0.32 & 0.46$\pm$0.32 & 9.92$\pm$5.12 & 6.75$\pm$3.08 & 3.27$\pm$1.64 & 5.03$\pm$3.82 & 6.18$\pm$4.62\\
    \bottomrule
  \end{tabular}
  }
  \label{tab:sw benchmark}
\end{table}
\begin{table*}[tb!]
  \caption{\textbf{Performance of Think2Drive on \yjc{the 39 scenarios in CARLA v2.}} The success rate denotes the statistical frequency of achieving 100\% route completion with zero instances of infraction. A high success rate does not necessarily mean a high driving score, e.g. in YieldToEmergencyVehicle, the vehicle may finish its route but fails to yield to the emergency vehicle. }
   \resizebox{1.0\textwidth}{!}{
  \begin{tabular}{@{}cccccccc@{}}
    \toprule
    Scenario & \makecell{Success Rate}  &Scenario & \makecell{Success Rate} & Scenario & \makecell{Success Rate}  &Scenario & \makecell{Success Rate}\\
    \midrule
     {ParkingExit} & 0.89 &{\makecell{Hazard\\AtSidelane}}& 0.75&{\makecell{Vinilla\\Turn}}& 0.99&{\makecell{Invading\\Turn}}& 0.90\\
    \midrule
     {\makecell{Signalized\\LeftTurn}} & 0.95 &{\makecell{Signalized\\RightTurn}}& 0.76 &{\makecell{OppositeVehicle\\TakingPriority}} & 0.89 &{\makecell{OppositeVehicle\\RunningRedLight}}& 0.85\\
     \midrule
     {\makecell{Accident}} & 0.81 &{\makecell{Accident\\TwoWays}}& 0.61 & {\makecell{Crossing\\BicycleFlow}} & 0.83 &{\makecell{Highway\\CutIn}}& 1.0\\
    \midrule
     {\makecell{Construction}} &0.84 &{\makecell{Construction\\TwoWays}}&0.72& {\makecell{Interurban\\ActorFlow}} & 0.83 &{\makecell{InterurbanAdvanced\\ActorFlow}}& 0.8\\
    \midrule
     {\makecell{Blocked\\Intersection}} &0.80 &{\makecell{Enter\\ActorFlow}}& 0.65& {\makecell{NonSignalized\\RightTurn}} & 0.75 &{\makecell{NonSignalizedJunction\\LeftTurnEnterFlow}}& 0.67\\
     \midrule
     {\makecell{MergerInto\\SlowTraffic}} &0.67 &{\makecell{MergerInto\\SlowTrafficV2}}& 0.87& {\makecell{Highway\\Exit}} & 0.83 &{\makecell{NonSignalized\\JunctionLeftTurn}}& 0.79\\
    \midrule
     {\makecell{SignalizedJunction\\LeftTurnEnterFlow}} &0.86 &{\makecell{Vehicle\\TurningRoute}}& 0.78& {\makecell{VehicleTurning\\RoutePedestrian}} & 0.75 &{\makecell{Pedestrain\\Crossing}}& 0.91\\
     \midrule
     {\makecell{YieldTo\\EmergencyVehicle}} &0.92 &{\makecell{Hard\\Brake}}& 1.0& {\makecell{Parking\\CrossingPedestrian}} & 0.98 &{\makecell{Dynamic\\ObjectCrossing}}& 0.94\\
     \midrule
     {\makecell{Vehicles\\DooropenTwoWays}} &0.78 &{\makecell{HazardAt\\SideLaneTwoWays}}& 0.92& {\makecell{Parked\\Obstacle}} & 0.90 &{\makecell{ParkedObstacle\\TwoWays}}& 0.91\\
     \midrule
     {\makecell{Static\\CutIn}} &0.85 &{\makecell{Parking\\CutIn}}& 0.90& {\makecell{ControlLoss}} & 0.78 \\
    \bottomrule
  \end{tabular}}
  \label{tab: overall scenario success rate}
\end{table*}

\section{Experiment}
\label{sec:experiment}
\subsection{CARLA Leaderboard v2}
\label{subsec:leaderboardv2}
\lqf{CARLA v2 is based on the CARLA simulator with version bigger than 0.9.13 (V1 on 0.9.10).} 
We evaluate our planner with CARLA 0.9.14. CARLA team initially proposed Leaderboard v1, which is composed of basic tasks such as lane following, turning, collision avoidance, and etc. Then, to facilitate quasi-realistic urban driving, CARLA v2 is released, which encompasses multitude complex scenarios previously absent in v1. These scenarios pose serious challenges. 

Since the release of CARLA v2, no team has managed to get a spot to tackle these scenarios, despite the availability of perfect logs scoring 100\% on each scenario, provided by the CARLA official platform to aid in related research. 
In our analysis, there are four primary reasons to the difficulty of v2: 
1) Extended Route Lengths: 
In CARLA v2, the routes extend between 7 to 10 kilometers, 
a substantial increase from the roughly 1-kilometer routes in v1. 
2) Complex and Abundant Scenarios: 
Each route contains around 60 scenarios, which require the driving methods to be able to handle complex road conditions and conduct subtle control.
3) Exponential decay scoring rules: 
The leaderboard employs a scoring mechanism that penalizes infractions through multiplication penalty factors \(<1\). 
In scenarios with extended routes and a multitude of scenarios, models struggle to attain high scores.  
4) Limited data: the CARLA team only provides a set of 90 training routes coupled with scenarios while routes randomly generated by researchers, does not have official API support for the placement of scenarios. 

\subsection{\benchmarkname\ Benchmark}
\label{subsec:CornerCaseRepo}
In the official benchmark, multiple scenarios are along a single long route, making it hard to train and evaluate the model.
To address this deficiency, we introduce the \benchmarkname\ benchmark, consisting of 1,600 routes for training and 390 routes for evaluation. Every route in the benchmark contains only one type of scenario with a length $<$ 300 meters so that the training and evaluation of different scenarios are decoupled. In the training set, there are 40 routes for each scenario and 40 routes without any scenarios. The routes are sampled randomly during the RL training process. 
There are 10 routes for each scenario in the evaluation routes. For evaluation, the routes are sampled sequentially until all routes have been evaluated. 
\benchmarkname\ supports the use of the CARLA metrics (e.g. driving scores, route completion) to analyze the performance of each scenario separately, providing convenience for debugging. 
\subsection{Weighted Driving Score}
\label{subsec:weighted_driving_score}
As described in~\cref{subsec:leaderboardv2}, the scoring rules of CARLA leaderboard is imperfect for driving policy evaluation. 
For instance, consider a driving model with an average infraction rate of 0.2 per kilometer and a penalty factor of 0.8. 
Under the hypothetical ideal condition where route completion is 100\% for both 5-kilometer and 10-kilometer test routes, the driving scores would be 0.8 and 0.64,  
i.e., \textbf{the longer the distance traveled, the lower the final driving score}. 
To avoid such counter-intuitive phenomenon, we propose a new metric named \textbf{Weighted Driving Score (WDS)}, formed as:
\begin{equation}
    \label{eq:metric}
    \begin{aligned}
    \text{WDS} = \text{RC}*\prod_{i}^{m} {\text{penalty}}_{i}^{n_i}
    \end{aligned}
\end{equation}
where $\text{RC}$ means route completion rate, $m$ is the total number of types of infractions considered, $\text{penalty}_{i}$ is the penalty factor for infraction type $i$ officially defined in CARLA,  and $n_i = \frac{\text{Number of Infractions}}{\text{Scenario Density}}$ (when there is no scenario, we set $n_i = \text{Number of Infractions}$ in which case Weighted Driving Score=Driving Score). Weighted Driving Score effectively balances the weight between route completion, number of infractions, and scenario density, providing a measure of the average infractions encountered by the ego vehicle over routes. 

\begin{table}[!tb]
\centering
\caption{\textbf{Performance on official test routes.}}
\begin{tabular}{l|l|ccc}
\toprule
Method                         & benchmark   & \makecell{Driving\\Scores} & \makecell{Weighted\\Driving Score} & \makecell{Route\\Complete \%} \\ \midrule
Roach (Expert)    & \multirow{2}{*}{CARLA Leaderboard v1}       & 84.0 & -- & 95.0 \\ 
Think2Drive (Ours)  &      & \textbf{90.2} & \textbf{90.2} & \textbf{99.7}  \\
\midrule
PPO (Expert) & \multirow{2}{*}{CARLA Leaderboard v2}       & 0.7 & 0.6 & 1.0 \\
Think2Drive (Ours)    &       & \textbf{56.8} & \textbf{91.7} & \textbf{98.6} \\ \bottomrule
\end{tabular}
\label{tab:test}
\end{table}

\subsection{Performance}
\label{subsec:performance}
\begin{figure}[tb!]
    \centering
    \begin{subfigure}{0.37\linewidth}
        \includegraphics[width=\linewidth]{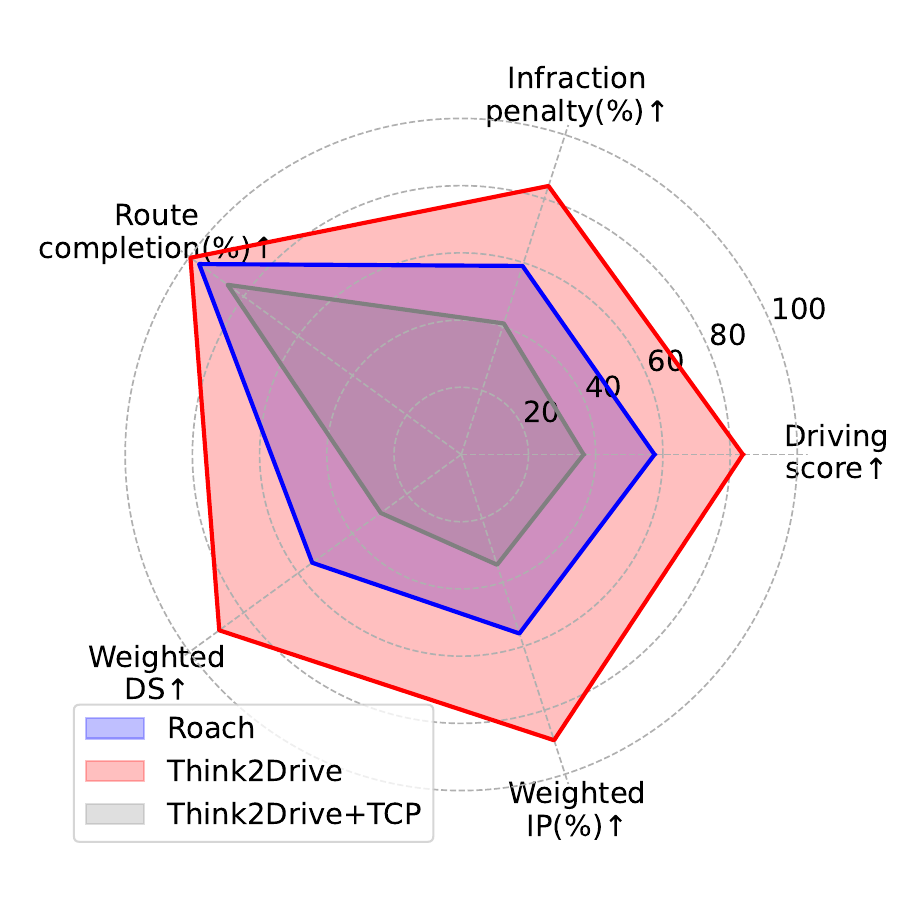}
        \caption{Driving performance.}
      \end{subfigure}
        \rulesep
      \begin{subfigure}{0.41\linewidth}
          \includegraphics[width=\linewidth]{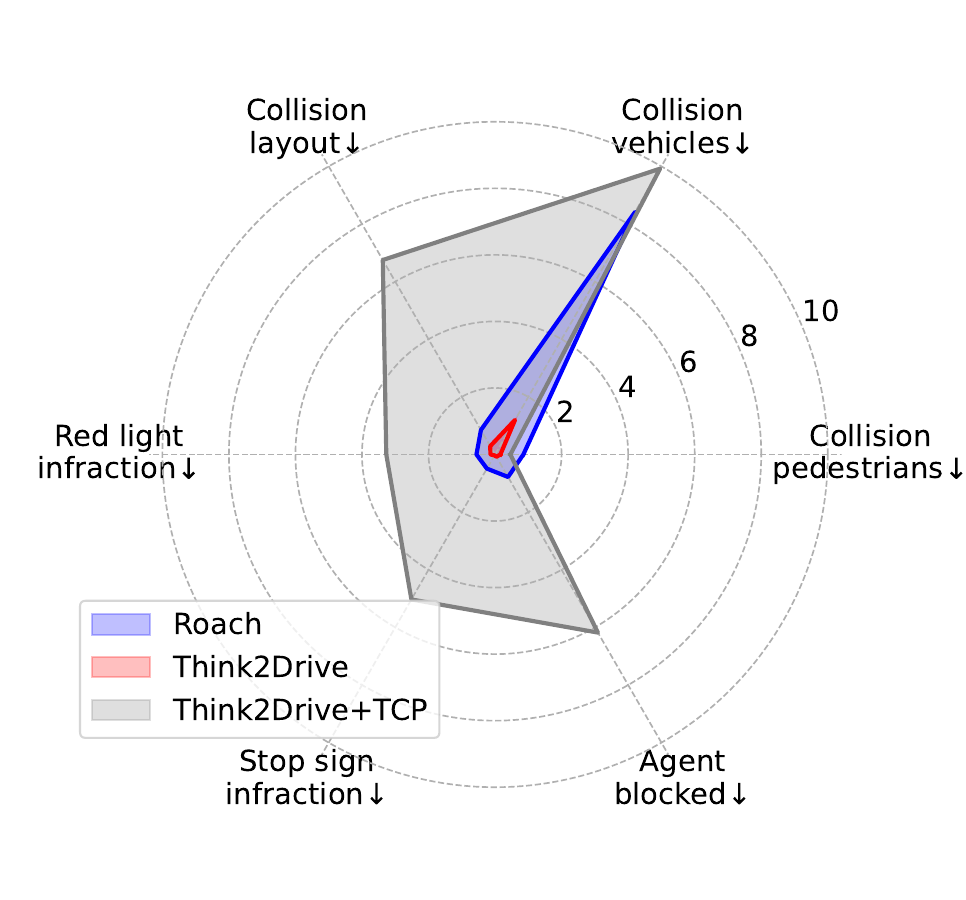}
          \caption{Infractions per kilometer.}
      \end{subfigure}
    \caption{\textbf{Driving performance and infractions on \benchmarkname}. 
    }
    \label{fig:radar}
\end{figure}

\cref{tab:sw benchmark} and \cref{fig:radar} show results on \benchmarkname\. The overall training time on one A6000 GPU with AMD Epyc 7542 CPU -- 128 logical cores is 3 days. For the baseline expert model, we implement Roach~\cite{zhang2021roach}, where we replace our model-based RL model with model-free PPO~\cite{schulman2017proximal} and keep all other techniques the same. Both experts are trained on 1600 routes and evaluated on other 390 routes of \benchmarkname\ benchmark for 3 runs. Think2Drive outperforms Roach by a large margin, showing the advantages of model-based RL.

We also choose and train an end-to-end baseline  TCP~\cite{wu2022trajectory}, a lightweight yet competitive student model  on CARLA Leaderboard v1, as the imitation learning agent. We train TCP with 200K frames collected by the Think2Drive expert under different weather. We could observe that TCP, as a student model with only raw sensor inputs, has a large performance gap with both expert models, as caused by the difficulty of perception as well as the imitation process. 

We also evaluate Think2Drive on the official test routes of CARLA Leaderboard v1 \& v2 and compare it with the expert model of Roach. As illustrated in \cref{tab:test}, Think2Drive not only outperforms Roach in the easier CARLA Leaderboard v1(no scenarios) but also achieves significantly superior performance in the more complex v2 routes. 
PPO achieved remarkably low scores, a consequence of its rapid convergence to local optima, beyond which the policy ceases to improve further even with the help of the reset technique. 
We argue that the reason for this phenomenon is that after each reset, PPO has to relearn the policy from the trajectories stored in the replay buffer, which contain inherent reward noise due to the AD characteristics. 
Conversely, a model-based planner can get accurate and smooth rewards from the world model.

\subsection{Infraction Analysis on Hard Scenarios}
We analyze the performance of Think2Drive in all scenarios, and give the success rate of the scenarios in \cref{tab: overall scenario success rate}. 

The scenarios \textit{Accident}, \textit{Construction}, and \textit{HazardAtSidelane} along with their respective \textit{TwoWays} versions, belong to the category of \textit{RouteObstacles} scenarios. 
In these scenarios, the ego vehicle is required to perform lane changes to maneuver around obstacles, particularly in the case of the \textit{TwoWays} versions where the ego vehicle needs to switch to the opposite lane. 
Such scenarios demand the ego vehicle to acquire a sophisticated lane negotiation policy, especially in the \textit{TwoWays} scenarios where the ego vehicle must execute lane changing, maneuver around obstacles, and return to its original lane within a short time window. 
Failed cases in these scenarios typically result from collisions with an opposite car during the process of returning to the original lane after bypassing the obstacles. 
CARLA Leaderboard v2 generates randomly the opposite traffic flow with speed and interval range within $[8,18]$ and $[15,50]$ (typical value, may vary with specific road conditions) in the \textit{TwoWays} scenarios, which may lead to a significantly constrained time window (e.g.less 1 second) for bypassing obstacles when the speed is large while the interval is small. Consequently, the ego vehicle is required to rapidly accelerate from $speed=0$ to its maximum speed, and the ego vehicle usually runs at a high speed and is close to the opposite car when returning to its original lane, which leads to a high risk of collisions. 

The scenarios \textit{SignalizedLeftTurn}, \textit{CrossingBicycleFlow}, \textit{SignalizedRightTurn} and \textit{BlockedInterSection} belong to \textit{JunctionNegotiate} type. 
In these scenarios, the ego vehicle has to interrupt the opposite dense car or bicycle flow, merge into the dense traffic flow, and stop at the junction to await road clearance. In CARLA Leaderboard v2, the traffic flow of these scenarios is configured to be very aggressive, meaning it does not proactively yield to the ego vehicle. 
The ego vehicle needs to maintain a reasonable distance from other vehicles to avoid collisions. 
For instance, in the \textit{SignalizedRightTurn} scenario, it is expected to merge into traffic with an interval within $[15,25]$ meters and a speed within $[12,20]$ m/s. With a vehicle length of approximately 3 meters, the ego vehicle must not only accelerate rapidly to match the traffic speed in a short time but also  maintain a safe following distance from other vehicles. 


\begin{figure*}[tb!]
    \centering
    \includegraphics[width=0.98\linewidth]{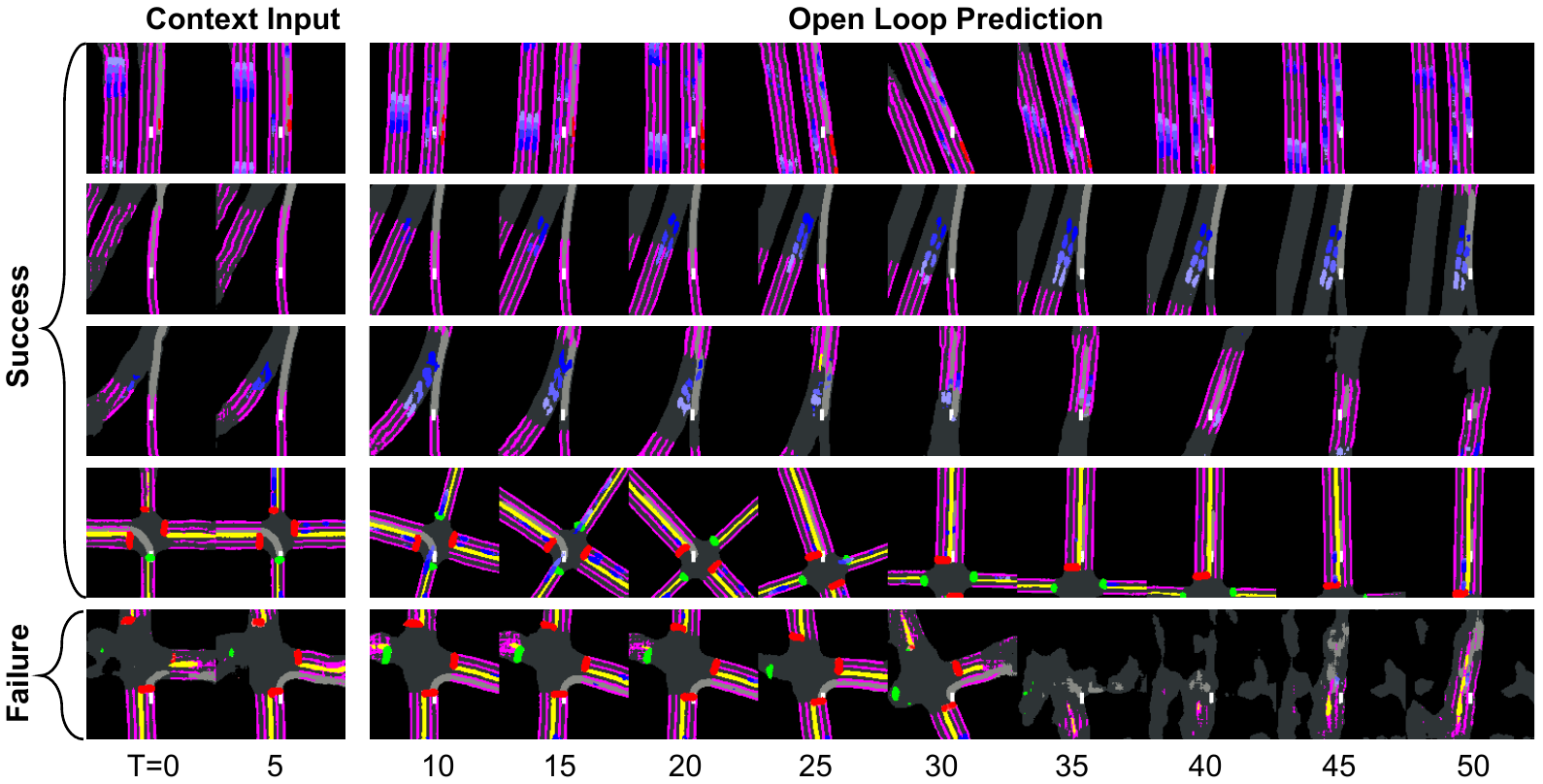}
    \caption{\textbf{Prediction by the world model using the first 5 frames.} It can predict reasonable future frames. In the failure case, the planner runs a red light and the world model terminates the episode, so the subsequent predictions are randomly generated.
    }
    \label{fig:prediction}
\end{figure*}

\subsection{Visualization of World Model Prediction}
The world model is capable of imaging observation transitions and future rewards based on the agent's actions, 
and it can decode them back into the interpretable masks under BEV.  \cref{fig:prediction} visualizes the initial input and the predicted BEV masks within timestep 50. We could observe that the world model could generate authentic future states, demonstrating one advantage of adopting model-based RL for AD - the transition function is usually easy to learn.

\begin{figure}[tb!]
    \centering
    \includegraphics[width=0.7\linewidth]{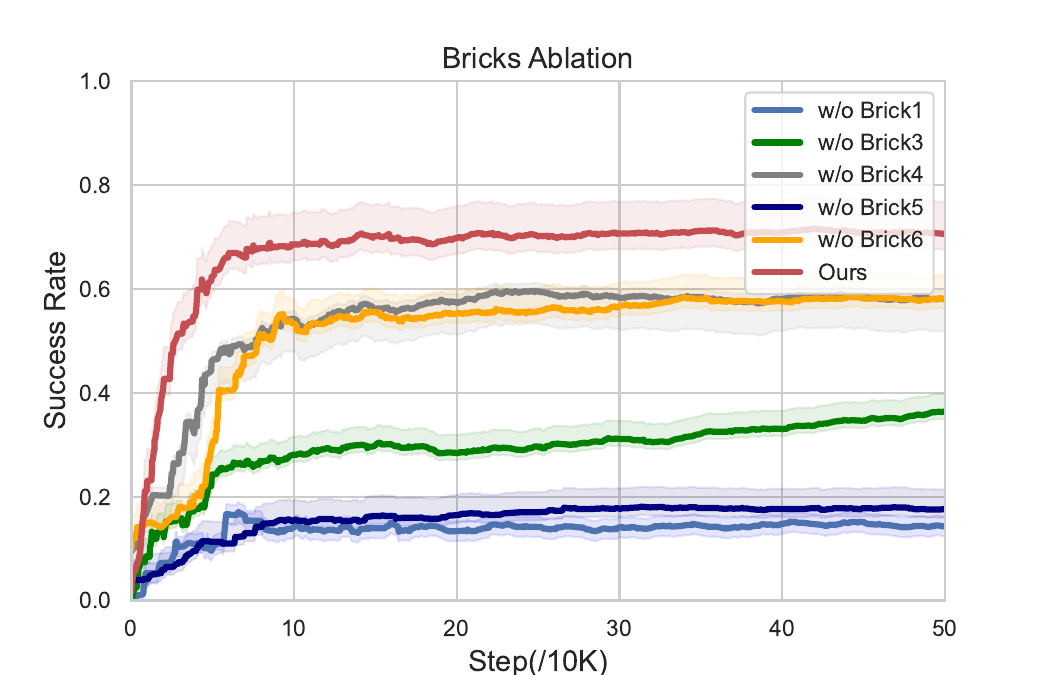}
    \caption{\textbf{Ablation of different bricks devised in the paper.}}
    \label{fig:brick_ablation}
\end{figure}
\subsection{Ablation Study}
\label{subsec:ablation}
We conduct ablation on bricks 1, 3-6 (bricks 2 and 7 are foundational for Think2Drive).   \cref{fig:brick_ablation} presents the results over 500K steps, showing that the absence of any single brick significantly diminishes the performance of Think2Drive.  Specifically, bricks 1 and 5 exert the most substantial impact on the final performance. The omission of the warmup stage (brick 5) results in an overly steep learning curve, while forgoing the reset technique (brick 1) predisposes the planner to be stuck in policies only effective in some easy scenarios, both of which lead to the model being trapped in the local optima. The absence of priority sampling (brick 3) is observed to reduce the model's exploration efficiency, evidenced by the ascending yet slow curve (w/o brick 3). Brick 6 affects the learning efficiency of the planner, where employing a higher training ratio for the planner enables the model to achieve superior performance within the same number of steps. The absence of a steering cost function (brick 4) compromises vehicle steering stability, increasing the propensity for collisions.

\section{Implement Details}
\label{sec:implement_details}
For the input representation, we utilize BEV semantic segmentation masks $i_{RL} \in \{0, 1\}^{H \times W \times C}$ as image input, where each channel denotes the occurrence of certain types of objects. It is generated from the privileged information obtained from the simulator and consists of $C$ masks of size $H \times W$. 
In these $C$ masks, the route, lanes, and lane markings are all static and thus could be represented by a single mask while those dynamic objects (e.g. vehicles and pedestrians) have $T$ masks where each mask represents their state at one history time-step.
Additionally, we feed speed, control action, and relative height of the ego vehicle at the previous time steps as input $v_{RL} \in \mathbb{R}^{K}$. 
More specific details in \cref{appendix:IO}, \cref{tab:bev} and \cref{tab:representation}. 

For output representation, we discretize the continuous action into 30 actions to reduce the complexity. 
The discretized actions are presented in \cref{tab:actions}. 
 
\noindent\textbf{Reward Shaping.}
It is shaped to make the planner keep safe driving and finish the route as much as possible, which consists of four parts: 
\textbf{1)} Speed reward $r_{speed}$ is used to train the ego vehicle to keep a safe speed, depending on the distance to other objects and their type. 
\textbf{2)} Travel reward $r_{travel}$ is the distance traveled along the target routes at each tick of CARLA. 
Travel reward encourages the ego vehicle to finish more target routes. 
\textbf{3)} Deviation penalty $p_{deviation}$ is the negative value of the distance between the ego vehicle and the lane center. 
It is normalized by the max deviation threshold $D_{max}$. 
\textbf{4)} Steering cost $c_{steer}$ is used to make the ego vehicle drive more smoother. 
We set it as the difference between the current steer and the last one. 
The overall reward is given by:
\begin{equation}
     \begin{aligned}
        r = r_{speed} + \alpha_{tr} r_{travel} + \alpha_{de} p_{deviation} + \alpha_{st} c_{steer}
     \end{aligned}
\end{equation}


\section{Conclusion}
We have proposed a purely learning-based planner, Think2Drive, for quasi-realistic traffic scenarios. 
Benefiting from the model-based RL paradigm, It can drive proficiently in CARLA Leaderboard v2 with all 39 scenarios within 3 days of training on a single GPU. 
We also devise tailored bricks such as resetting
technique, automated scenario generation, termination-priority replay strategy, and steering cost function to address the obstacles associated with applying model-based RL to autonomous driving tasks. 
Think2Drive highlights and validates a feasible approach, model-based RL, for quasi-realistic autonomous driving. 
Our model can also serve as a data collection model, providing expert driving data for end-to-end autonomous driving models.





%
%
\newpage
\bibliographystyle{splncs04}
\bibliography{main}
\clearpage

\appendix
\begin{center}
    \Large \textbf{Think2Drive: Efficient Reinforcement Learning by Thinking with Latent World Model for Autonomous Driving (in CARLA-v2)} \\ 
    \vspace{-3mm}
\end{center}
\section{Discription of CARLA v2 Scenarios }
CARLA v2 provides 39 corner scenarios which are common in real driving environments. 
These scenarios can generally be divided into two categories: regular road scenarios and junction scenarios. 
Here we give a detailed description of each scenario: 
\subsection{Regular road scenarios}
These scenarios are related to regular road segments

\textbf{\textit{1. ControlLoss}} 

\begin{minipage}{0.3\textwidth}
\includegraphics[width=\textwidth]{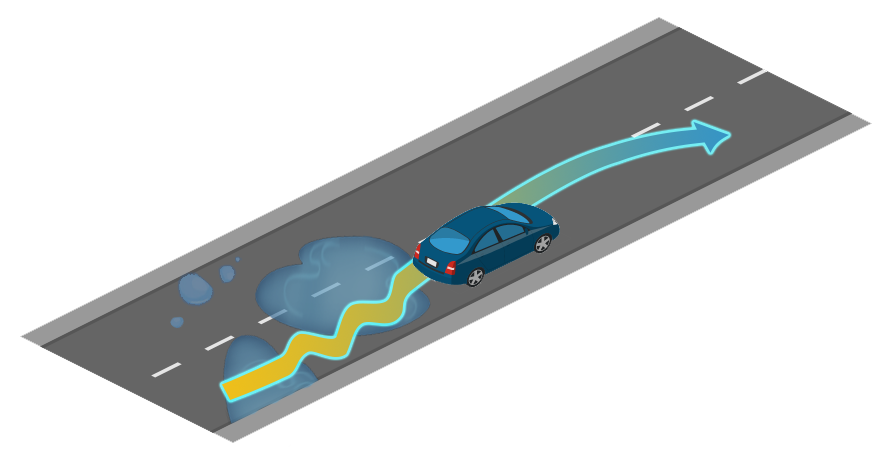}
\end{minipage}
\begin{minipage}{0.67\textwidth}
The ego vehicle loses control due to bad conditions on the road and it must recover, coming back to its original lane.
\end{minipage}

\textbf{\textit{2. ParkingExit}}

\begin{minipage}{0.3\textwidth}
\includegraphics[width=\textwidth]{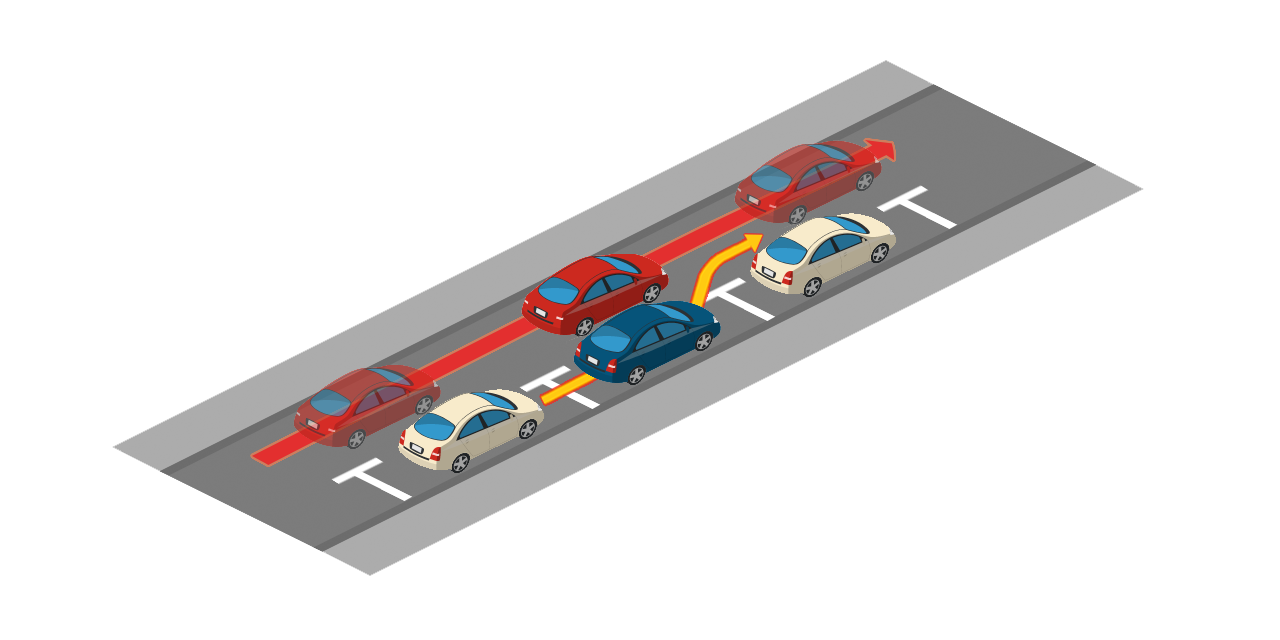}
\end{minipage}
\begin{minipage}{0.67\textwidth}
The ego vehicle must exit a parallel parking bay into a flow of traffic.
\end{minipage}

\textbf{\textit{3. ParkingCutIn}}

\begin{minipage}{0.3\textwidth}
\includegraphics[width=\textwidth]{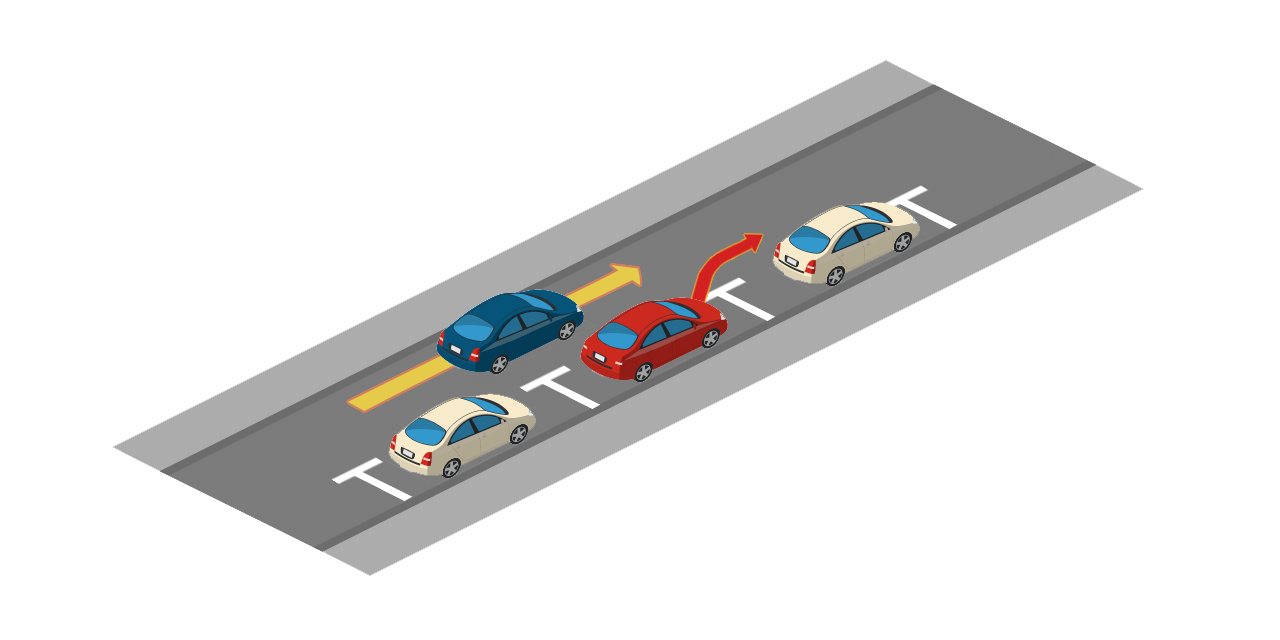}
\end{minipage}
\begin{minipage}{0.67\textwidth}
The ego vehicle must slow down or brake to allow a parked vehicle exiting a parallel parking bay to cut in front. 
\end{minipage}

\textbf{\textit{4. StaticCutIn}}

\begin{minipage}{0.3\textwidth}
\includegraphics[width=\textwidth]{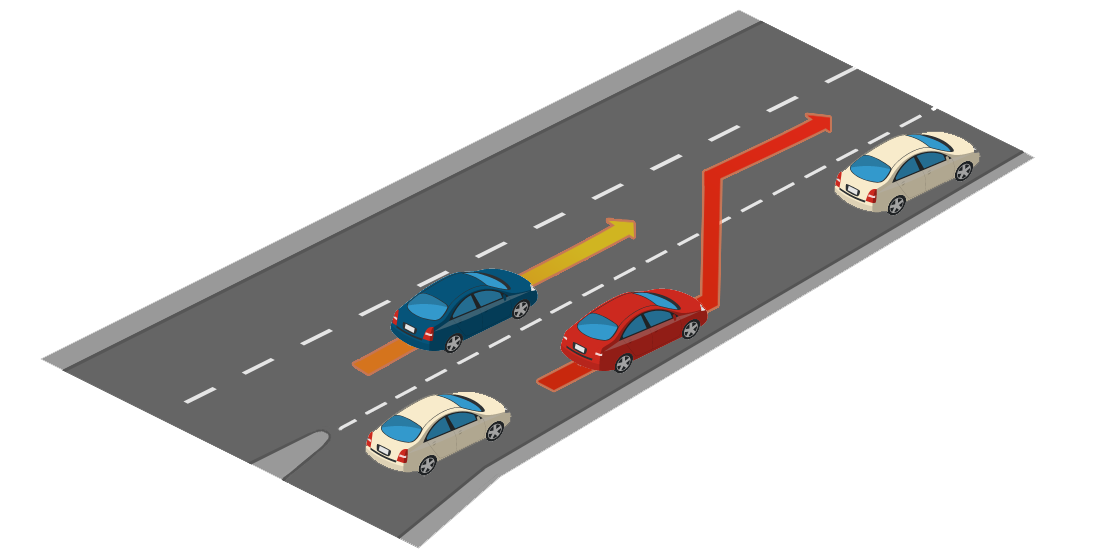}
\end{minipage}
\begin{minipage}{0.67\textwidth}
The ego vehicle must slow down or brake to allow a vehicle of the slow traffic flow in the adjacent lane to cut in front. 
Compared to \textit{ParkingCutIn}, there are more cars in the adjacent lane and any one of them may cut in. 
\end{minipage}

\textbf{\textit{5. ParkedObstacle}}

\begin{minipage}{0.3\textwidth}
\hspace{0.8cm}\includegraphics[width=0.5\textwidth]{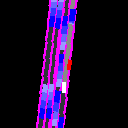}
\end{minipage}
\begin{minipage}{0.67\textwidth}
The ego vehicle encounters a parked vehicle blocking part of the lane and must perform a lane change into traffic moving in the same direction to avoid it. 
\end{minipage}

\textbf{\textit{6. ParkedObstacleTwoWays}}

\begin{minipage}{0.3\textwidth}
\hspace{0.8cm}\includegraphics[width=0.5\textwidth]{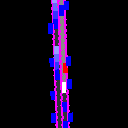}
\end{minipage}
\begin{minipage}{0.67\textwidth}
The '\textit{TwoWays}' version of \textit{ParkedObstacle}. 
The ego vehicle encounters a parked vehicle blocking the lane and must perform a lane change into traffic moving in the opposite direction to avoid it. 
\end{minipage}

\textbf{\textit{7. Construction}}

\begin{minipage}{0.3\textwidth}
\includegraphics[width=\textwidth]{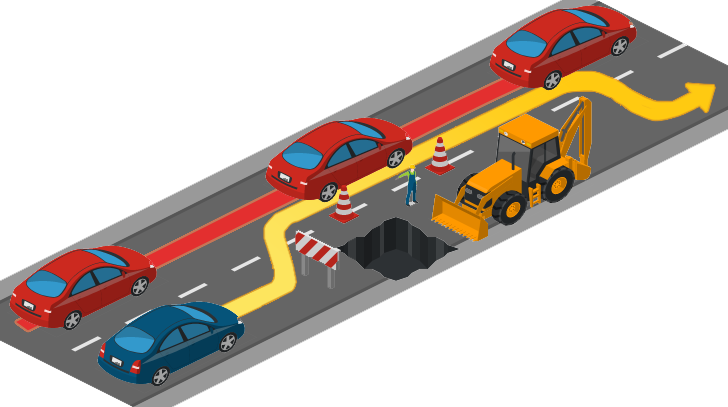}
\end{minipage}
\begin{minipage}{0.67\textwidth}
The ego vehicle encounters a construction site blocking and must perform a lane change into traffic moving in the same direction to avoid it. 
Compared to \textit{ParkedObstacle}, the construction occupies more width of the lane. 
The ego vehicle has to completely deviate from its task route temporarily to bypass the construction zone. 
\end{minipage}

\textbf{\textit{8. ConstructionTwoWays}}

\begin{minipage}{0.3\textwidth}
\includegraphics[width=\textwidth]{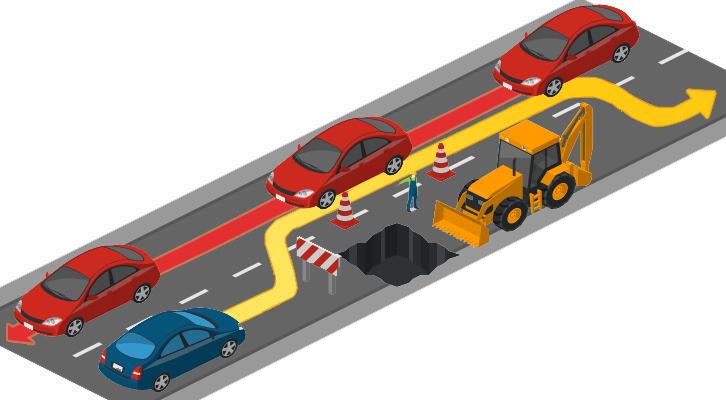}
\end{minipage}
\begin{minipage}{0.67\textwidth}
The '\textit{TwoWays}' version of \textit{Construction}. 
\end{minipage}

\textbf{\textit{9. Accident}} 

\begin{minipage}{0.3\textwidth}
\hspace{0.8cm}\includegraphics[width=0.5\textwidth]{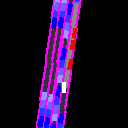}
\end{minipage}
\begin{minipage}{0.67\textwidth}
The ego vehicle encounters multiple accident cars blocking part of the lane and must perform a lane change into traffic moving in the same direction to avoid it. 
Compared to \textit{ParkedObstacle} and \textit{Construction}, these accident cars occupy more length along the lane. 
The ego vehicle has to completely deviate from its task route for a longer time to bypass the accident zone. 
\end{minipage}

\textbf{\textit{10. AccidentTwoWays}}

\begin{minipage}{0.3\textwidth}
\hspace{0.8cm}\includegraphics[width=0.5\textwidth]{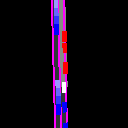}
\end{minipage}
\begin{minipage}{0.67\textwidth}
The '\textit{TwoWays}' version of \textit{Accident}. 
Compared to \textit{ParkedObstacleTwoWays} and \textit{ConstructionTwoWays}, there is a much shorter time window for the ego vehicle to bypass the route obstacles (i.g. accident cars). 
\end{minipage}

\textbf{\textit{11. HazardAtSideLane}}

\begin{minipage}{0.3\textwidth}
\includegraphics[width=\textwidth]{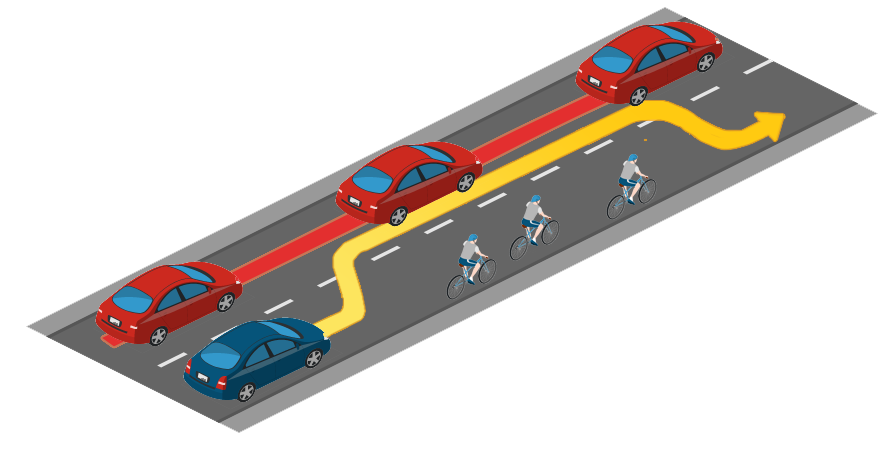}
\end{minipage}
\begin{minipage}{0.67\textwidth}
The ego vehicle encounters a slow-moving hazard blocking part of the lane. 
The ego vehicle must brake or maneuver next to a lane of traffic moving in the same direction to avoid it. 
\end{minipage}

\textbf{\textit{12. HazardAtSideLaneTwoWays}}

\begin{minipage}{0.3\textwidth}
\includegraphics[width=\textwidth]{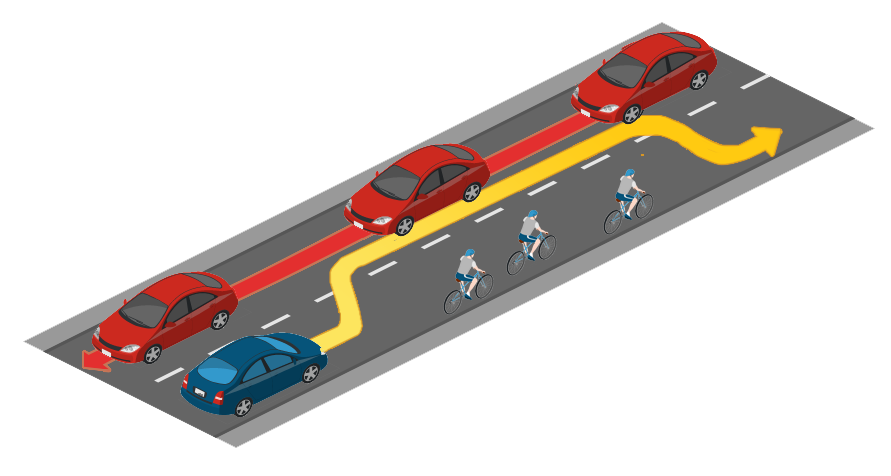}
\end{minipage}
\begin{minipage}{0.67\textwidth}
The ego vehicle encounters a slow-moving hazard blocking part of the lane. The ego vehicle must brake or maneuver to avoid it next to a lane of traffic moving in the opposite direction. 
\end{minipage}

\textbf{\textit{13. VehiclesDooropenTwoWays}}

\begin{minipage}{0.3\textwidth}
\includegraphics[width=\textwidth]{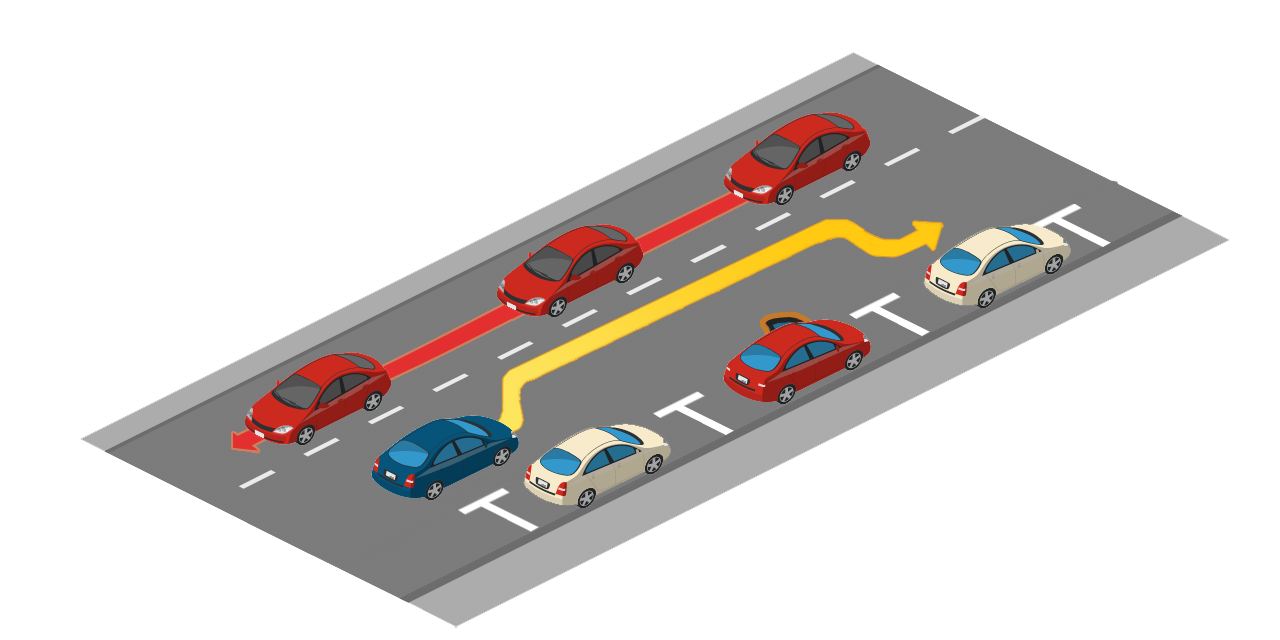}
\end{minipage}
\begin{minipage}{0.67\textwidth}
The ego vehicle encounters a parked vehicle opening a door into its lane and must maneuver to avoid it. 
\end{minipage}

\textbf{\textit{14. DynamicObjectCrossing}}

\begin{minipage}{0.3\textwidth}
\hspace{0.8cm}\includegraphics[width=0.5\textwidth]{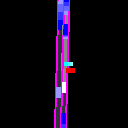}
\end{minipage}
\begin{minipage}{0.67\textwidth}
A walker or bicycle behind a static prop crosses the road suddenly when the ego vehicle is close to the prop. 
The ego vehicle must make a hard brake promptly. 
\end{minipage}

\textbf{\textit{15. ParkingCrossingPedestrian}}

\begin{minipage}{0.3\textwidth}
\includegraphics[width=\textwidth]{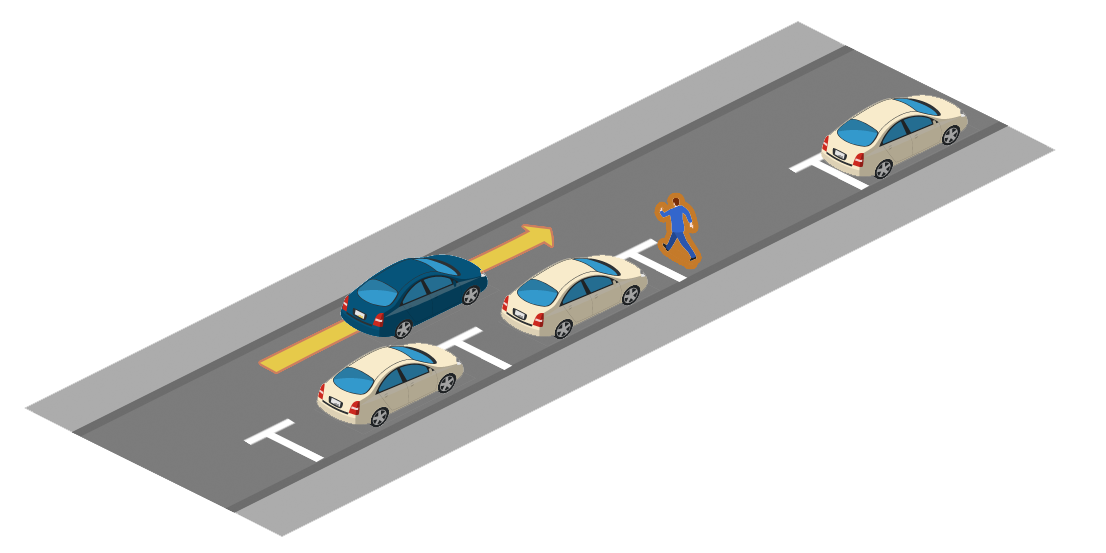}
\end{minipage}
\begin{minipage}{0.67\textwidth}
The ego vehicle encounters a pedestrian emerging from behind a parked vehicle and advancing into the lane. 
The ego vehicle must brake or maneuver to avoid it. 
Compared to \textit{DynamicObjectCrossing}, the pedestrian is closer to the road and the ego vehicle has to act more timely. 
\end{minipage}

\textbf{\textit{16. HardBrake}}

\begin{minipage}{0.3\textwidth}
\includegraphics[width=\textwidth]{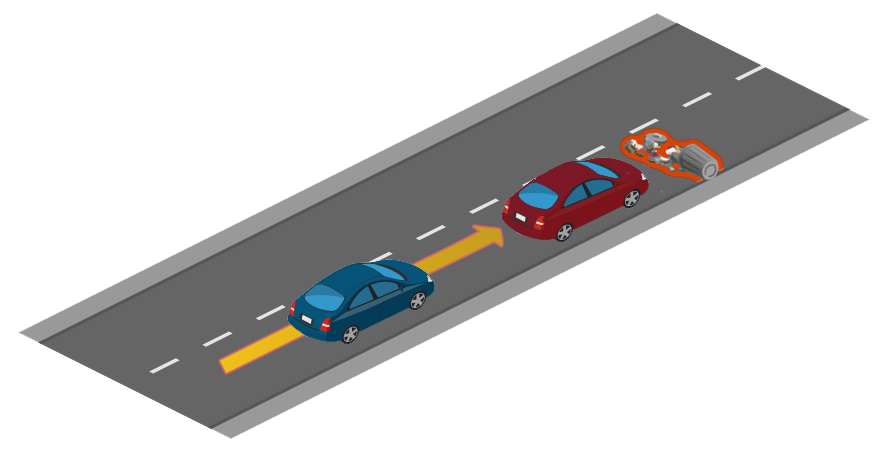}
\end{minipage}
\begin{minipage}{0.67\textwidth}
The leading vehicle decelerates suddenly and the ego vehicle must perform an emergency brake or an avoidance maneuver. 
\end{minipage}

\textbf{\textit{17. YieldToEmergencyVehicle}}

\begin{minipage}{0.3\textwidth}
\includegraphics[width=\textwidth]{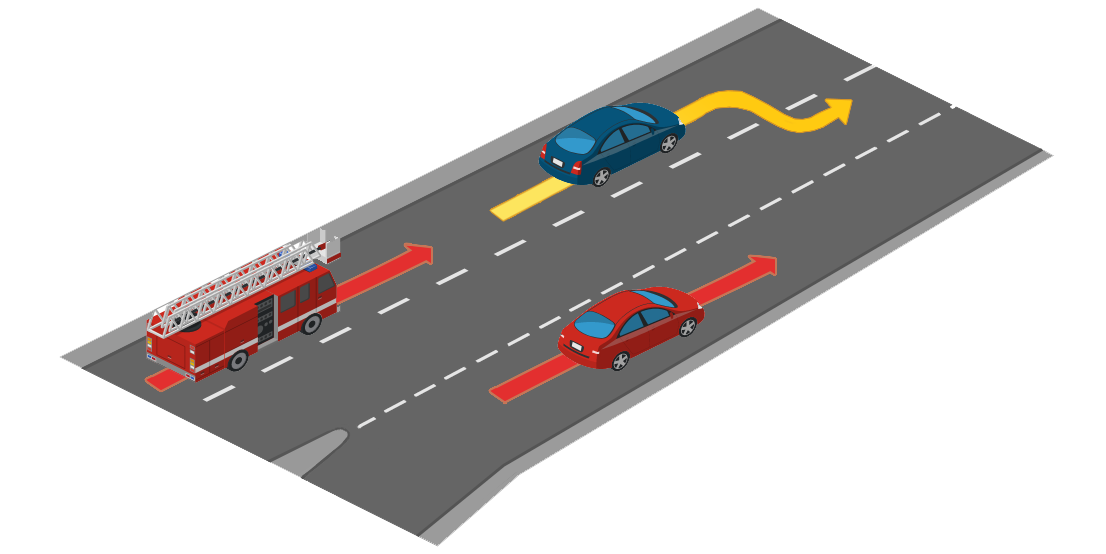}
\end{minipage}
\begin{minipage}{0.67\textwidth}
The ego vehicle is approached by an emergency vehicle coming from behind. 
The ego vehicle must maneuver to allow the emergency vehicle to pass.
\end{minipage}

\textbf{\textit{18. InvadingTurn}}

\begin{minipage}{0.3\textwidth}
\hspace{0.8cm}\includegraphics[width=0.5\textwidth]{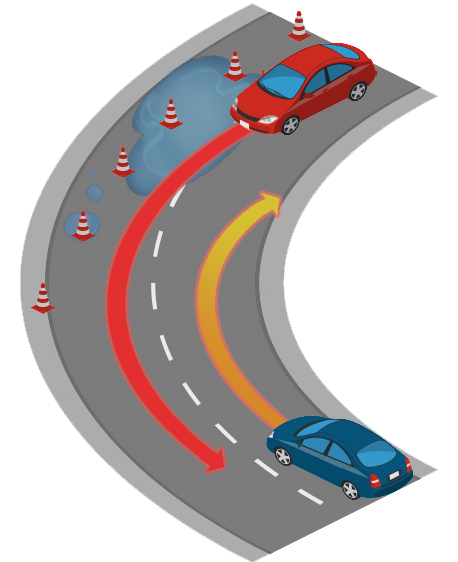}
\end{minipage}
\begin{minipage}{0.67\textwidth}
When the ego vehicle is about to turn right, a vehicle coming from the opposite lane invades the ego's lane, forcing the ego to move right to avoid a possible collision. 
\end{minipage}

\subsection{Junction scenarios}
These scenarios are related to junctions. 

\textbf{\textit{1. PedestrainCrossing}}

\begin{minipage}{0.3\textwidth}
\hspace{0.8cm}\includegraphics[width=0.5\textwidth]{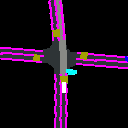}
\end{minipage}
\begin{minipage}{0.67\textwidth}
While the ego vehicle is entering a junction, a group of natural pedestrians suddenly cross the road and ignore the traffic light. 
The ego vehicle must stop and wait for all pedestrians to pass even though there is a green traffic light or a clear junction. 
\end{minipage}

\textbf{\textit{2. VehicleTurningRoutePedestrian}}

\begin{minipage}{0.3\textwidth}
\includegraphics[width=\textwidth]{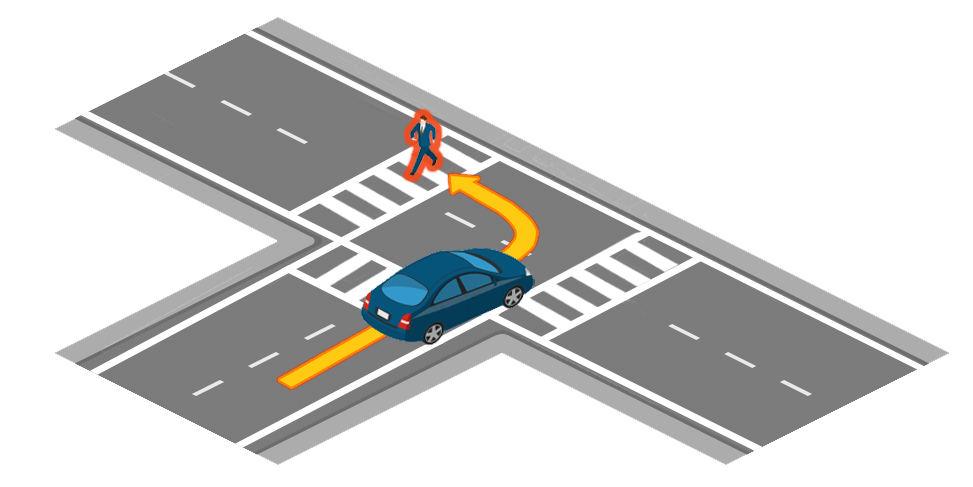}
\end{minipage}
\begin{minipage}{0.67\textwidth}
While performing a maneuver, the ego vehicle encounters a pedestrian crossing the road and must perform an emergency brake or an avoidance maneuver. 
\end{minipage}

\textbf{\textit{3. VehicleTurningRoute}}
While performing a maneuver, the ego vehicle encounters a bicycle crossing the road and must perform an emergency brake or an avoidance maneuver. 
Compared to \textit{VehicleTurningRoutePedestrian}, the bicycle moves faster and the ego has to brake earlier. 

\textbf{\textit{4. BlockedIntersection}}

\begin{minipage}{0.3\textwidth}
\includegraphics[width=\textwidth]{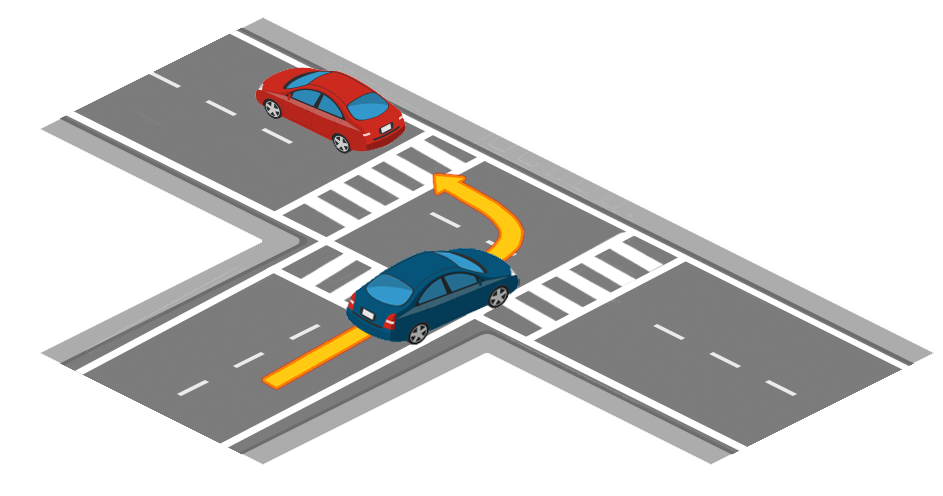}
\end{minipage}
\begin{minipage}{0.67\textwidth}
While performing a maneuver, the ego vehicle encounters a stopped vehicle on the road and must perform an emergency brake or an avoidance maneuver. 
\end{minipage}

\textbf{\textit{5. SignalizedJunctionLeftTurn}} 

\begin{minipage}{0.3\textwidth}
\includegraphics[width=\textwidth]{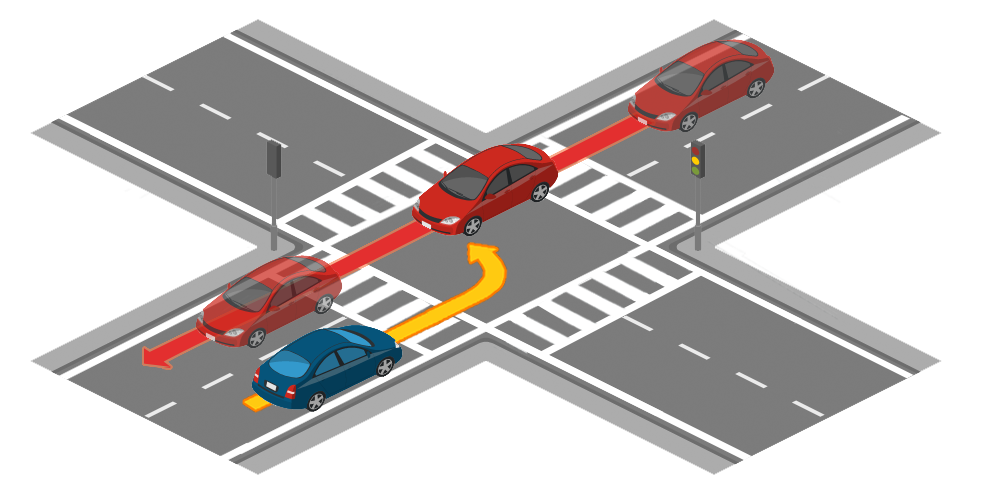}
\end{minipage}
\begin{minipage}{0.67\textwidth}
The ego vehicle is performing an unprotected left turn at an intersection, yielding to oncoming traffic. 
\end{minipage}

\textbf{\textit{6. SignalizedJunctionLeftTurnEnterFlow}} 

The ego vehicle is performing an unprotected left turn at an intersection, merging into opposite traffic. 

\textbf{\textit{7. NonSignalizedJunctionLeftTurn}} 

\begin{minipage}{0.3\textwidth}
\hspace{0.8cm}\includegraphics[width=0.5\textwidth]{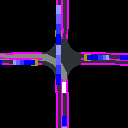}
\end{minipage}
\begin{minipage}{0.67\textwidth}
Non-signalized version of \textit{SignalizedJunctionLeftTurn}. 
The ego has to negotiate with the opposite vehicles without traffic lights. 
\end{minipage}

\textbf{\textit{8. NonSignalizedJunctionLeftTurnEnterFlow}} 

Non-signalized version of \textit{SignalizedJunctionLeftTurnEnterFlow}. 

\textbf{\textit{9. SignalizedJunctionRightTurn}} 

\begin{minipage}{0.3\textwidth}
\includegraphics[width=\textwidth]{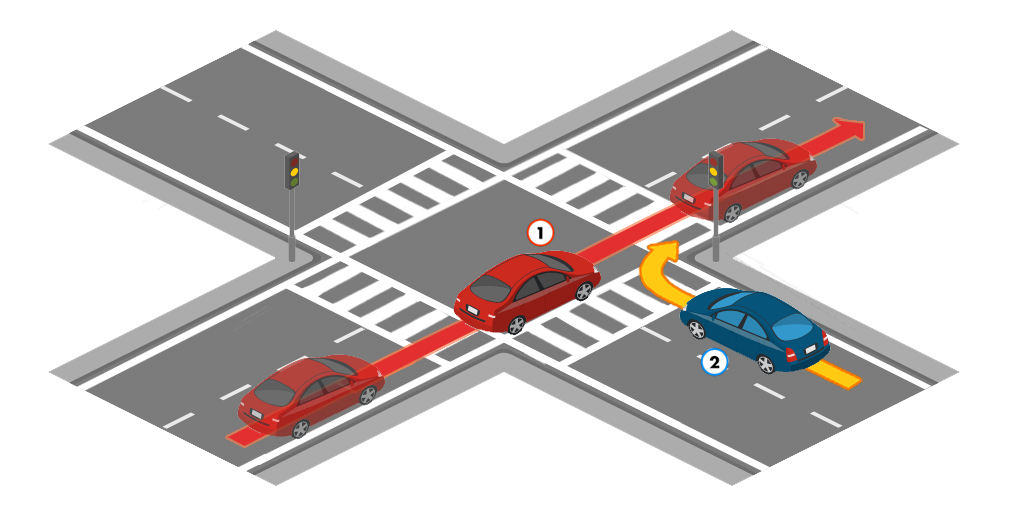}
\end{minipage}
\begin{minipage}{0.67\textwidth}
The ego vehicle is turning right at an intersection and has to safely merge into the traffic flow coming from its left. 
\end{minipage}

\textbf{\textit{10. NonSignalizedJunctionRightTurn}} 

Non-signalized version of \textit{SignalizedJunctionRightTurn}. 
The ego has to negotiate with the traffic flow without traffic lights.

\textbf{\textit{11. EnterActorFlows}} 

\begin{minipage}{0.3\textwidth}
\includegraphics[width=\textwidth]{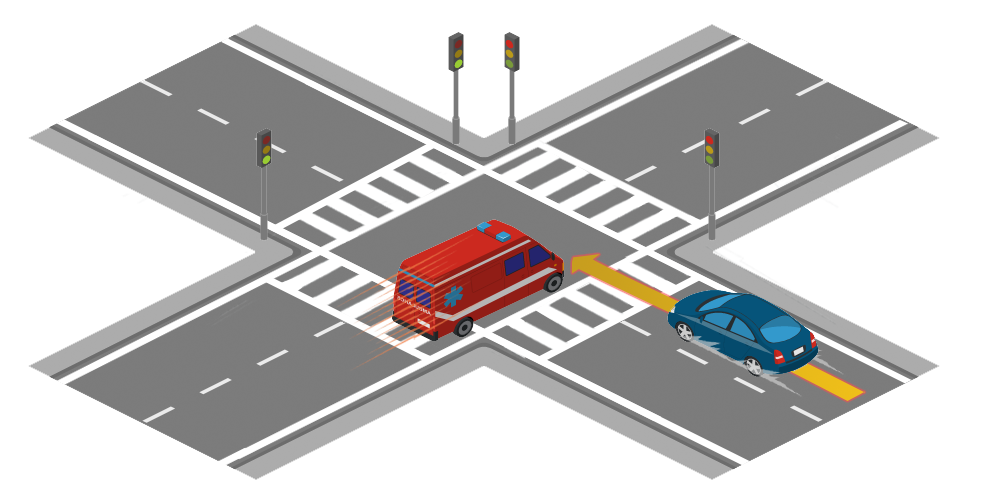}
\end{minipage}
\begin{minipage}{0.67\textwidth}
A flow of cars runs a red light in front of the ego when it enters the junction, forcing it to react (interrupting the flow or merging into the flow). 
These vehicles are 'special' ones such as police cars, ambulances, or firetrucks. 
\end{minipage}

\textbf{\textit{12. HighwayExit}} 

\begin{minipage}{0.3\textwidth}
\includegraphics[width=\textwidth]{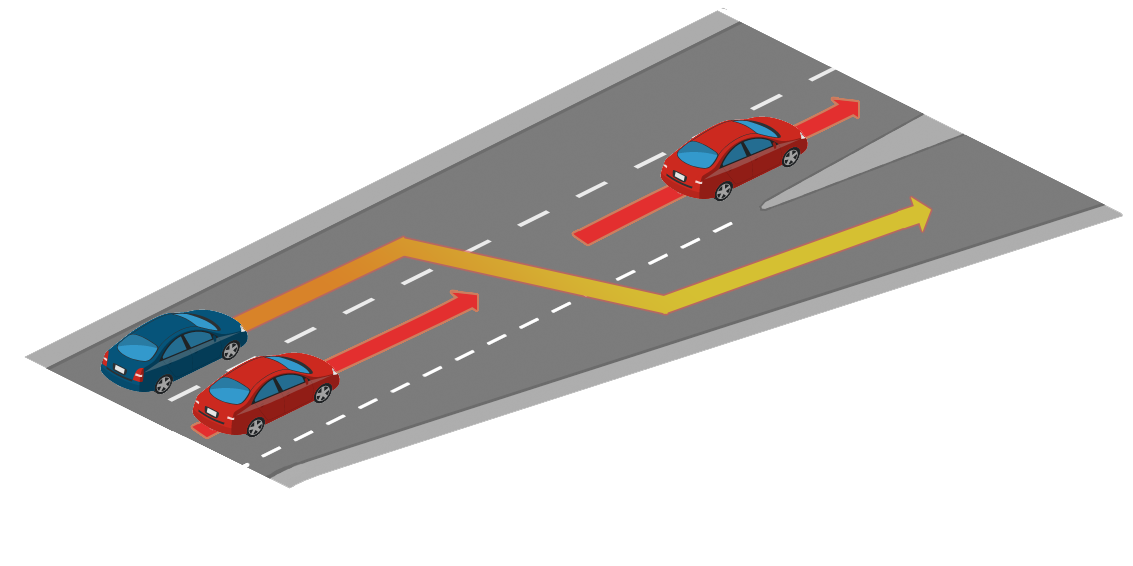}
\end{minipage}
\begin{minipage}{0.67\textwidth}
The ego vehicle must cross a lane of moving traffic to exit the highway at an off-ramp. 
\end{minipage}

\textbf{\textit{13. MergerIntoSlowTraffic}} 

\begin{minipage}{0.3\textwidth}
\hspace{0.8cm}\includegraphics[width=0.5\textwidth]{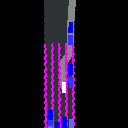}
\end{minipage}
\begin{minipage}{0.67\textwidth}
The ego vehicle must merge into a slow traffic flow on the off-ramp when exiting the highway. 
\end{minipage}

\textbf{\textit{14. MergerIntoSlowTrafficV2}} 

\begin{minipage}{0.3\textwidth}
\hspace{0.8cm}\includegraphics[width=0.5\textwidth]{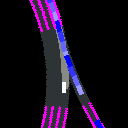}
\end{minipage}
\begin{minipage}{0.67\textwidth}
The ego vehicle must merge into a slow traffic flow coming from the on-ramp when driving on highway roads. 
\end{minipage}

\textbf{\textit{15. InterurbanActorFlow}} 

\begin{minipage}{0.3\textwidth}
\hspace{0.8cm}\includegraphics[width=0.5\textwidth]{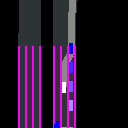}
\end{minipage}
\begin{minipage}{0.67\textwidth}
The ego vehicle leaves the interurban road by turning left, crossing a fast traffic flow. 
\end{minipage}

\textbf{\textit{16. InterurbanAdvancedActorFlow}} 

\begin{minipage}{0.3\textwidth}
\hspace{0.8cm}\includegraphics[width=0.5\textwidth]{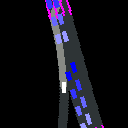}
\end{minipage}
\begin{minipage}{0.67\textwidth}
The ego vehicle incorporates into the interurban road by turning left, first crossing a fast traffic flow, and then merging into another one. 
\end{minipage}

\textbf{\textit{17. HighwayCutIn}} 

\begin{minipage}{0.3\textwidth}
\includegraphics[width=\textwidth]{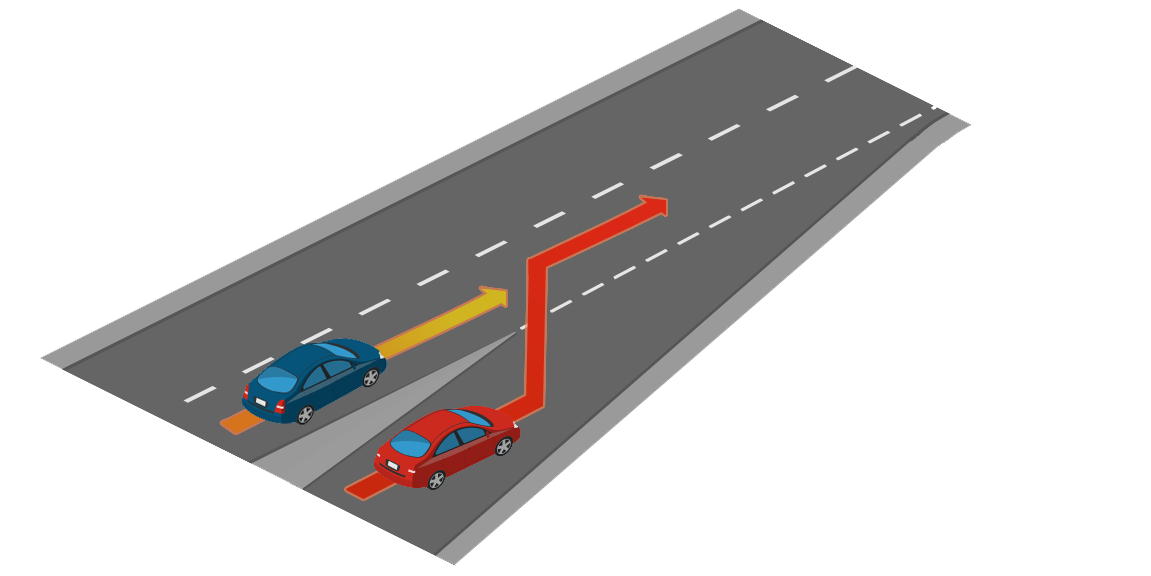}
\end{minipage}
\begin{minipage}{0.67\textwidth}
The ego vehicle encounters a vehicle merging into its lane from a highway on-ramp. 
The ego vehicle must decelerate, brake, or change lanes to avoid a collision. 
\end{minipage}

\textbf{\textit{18. CrossingBicycleFlow}} 

\begin{minipage}{0.3\textwidth}
\includegraphics[width=\textwidth]{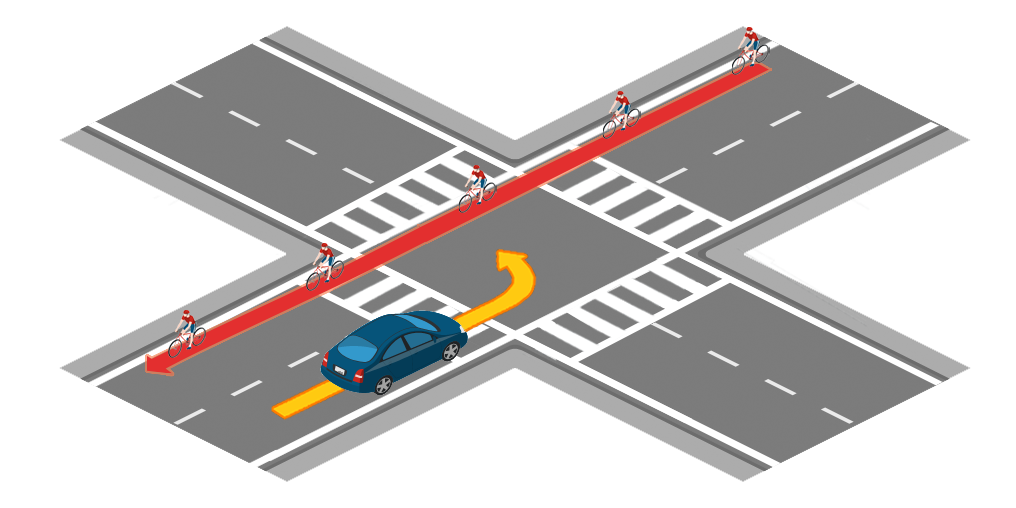}
\end{minipage}
\begin{minipage}{0.67\textwidth}
The ego vehicle needs to perform a turn at an intersection yielding to bicycles crossing from either the left. 
\end{minipage}

\textbf{\textit{19. OppositeVehicleRunningRedLight}} 

\begin{minipage}{0.3\textwidth}
\includegraphics[width=\textwidth]{pic/RunRedLight.png}
\end{minipage}
\begin{minipage}{0.67\textwidth}
The ego vehicle is going straight at an intersection but a crossing vehicle runs a red light, forcing the ego vehicle to avoid the collision.
\end{minipage}

\textbf{\textit{20. OppositeVehicleTakingPriority}} 

\begin{minipage}{0.3\textwidth}
\hspace{0.8cm}\includegraphics[width=0.5\textwidth]{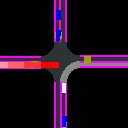}
\end{minipage}
\begin{minipage}{0.67\textwidth}
Non-signalized version of \textit{OppositeVehicleTakingPriority}. 
\end{minipage}

\textbf{\textit{21. VinillaTurn}}

\begin{minipage}{0.3\textwidth}
\hspace{0.8cm}\includegraphics[width=0.5\textwidth]{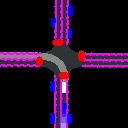}
\end{minipage}
\begin{minipage}{0.67\textwidth}
A basic scenario for the ego vehicle to learn the basic traffic rules, e.g. stop signs and traffic lights. 
\end{minipage}

\newpage
\section{Input \& Output Representation}
\label{appendix:IO}
\begin{table}[!th]
\caption{\textbf{Composition of the BEV representation.} Each static object occupies 1 dimension out of the $C=34$, while each dynamic object occupies 4 dimensions out of the $C=34$.}
\vspace{-6mm}
\begin{center}
\resizebox{1.0\textwidth}{!}{
\begin{tabular}{{cccccc|ccccccc}}
\hline
\multicolumn{13}{c}{$C=34$} \\
\hline
\multicolumn{6}{c|}{Static($C_s=1$)} & \multicolumn{7}{c}{Dynamic($C_d=4$)} \\
\hline
road & route & ego & lane & \makecell{yellow\\line} & \multicolumn{1}{c|}{\makecell{white\\line}} & vehicle & walker & \makecell{emergency\\car} & obstacle & \makecell{green\\ traffic light} & \makecell{yellow\&red\\traffic light} & \makecell{stop\\sign}\\
\hline
\end{tabular}}
\vspace{-6mm}
\label{tab:bev}
\end{center}
\end{table}

\vspace{-6mm}
\begin{table}[!th]
\caption{\textbf{Dimension of input and output representation.} The $T$ temporal masks are from historical time-steps: $[-16, -11, -6, -1]$.}
  \centering
  \begin{tabular}{cccccc}
    \toprule
    H & W & C & T & M \\
    \midrule
    128 & 128 & 34 & 4 & 30 \\
    \bottomrule
  \end{tabular}
  \vspace{-10pt}
   \label{tab:representation}
\end{table}

\begin{table}[!th]
  \centering
  \small
  \caption{\textbf{The discretized actions.} The continuous action space is decomposed into 30 discrete actions, each for specific values of throttle, steer, and brake. Each action is rational and legitimate.}
  \begin{tabular}{ccc|ccc|ccc}
     \toprule
     Throttle & Brake & Steer & Throttle & Brake & Steer & Throttle & Brake & Steer\\
    \midrule
     0 & 1 & 0 & 0.3 & 0 & {\footnotesize-}0.7 & 0.3 & 0 & 0.7\\
     0.7 & 0 & {\footnotesize-}0.5 & 0.3 & 0 & {\footnotesize-}0.5 & 0 & 0 & -1\\
     0.7 & 0 & {\footnotesize-}0.3 & 0.3 & 0 & {\footnotesize-}0.3 & 0 & 0 & {\footnotesize-}0.6\\
     0.7 & 0 & {\footnotesize-}0.2 & 0.3 & 0 & {\footnotesize-}0.2 & 0 & 0 & {\footnotesize-}0.3\\
     0.7 & 0 & {\footnotesize-}0.1 & 0.3 & 0 & {\footnotesize-}0.1 & 0 & 0 & {\footnotesize-}0.1\\
     0.7 & 0 & 0 & 0.3 & 0 & 0 & 0 & 0 & 0\\
     0.7 & 0 & 0.1 & 0.3 & 0 & 0.1 & 0 & 0 & 0.1\\
     0.7 & 0 & 0.2 & 0.3 & 0 & 0.2 & 0 & 0 & 0.3\\
     0.7 & 0 & 0.3 & 0.3 & 0 & 0.3 & 0 & 0 & 0.6\\
     0.7 & 0 & 0.5 & 0.3 & 0 & 0.5 & 0 & 0 & 1\\
    \bottomrule
  \end{tabular}
  \label{tab:actions}
\end{table}

\newpage
\section{Simulator Execution Mode}
\begin{figure}[h!]
    \centering  \includegraphics[width=0.90\linewidth]{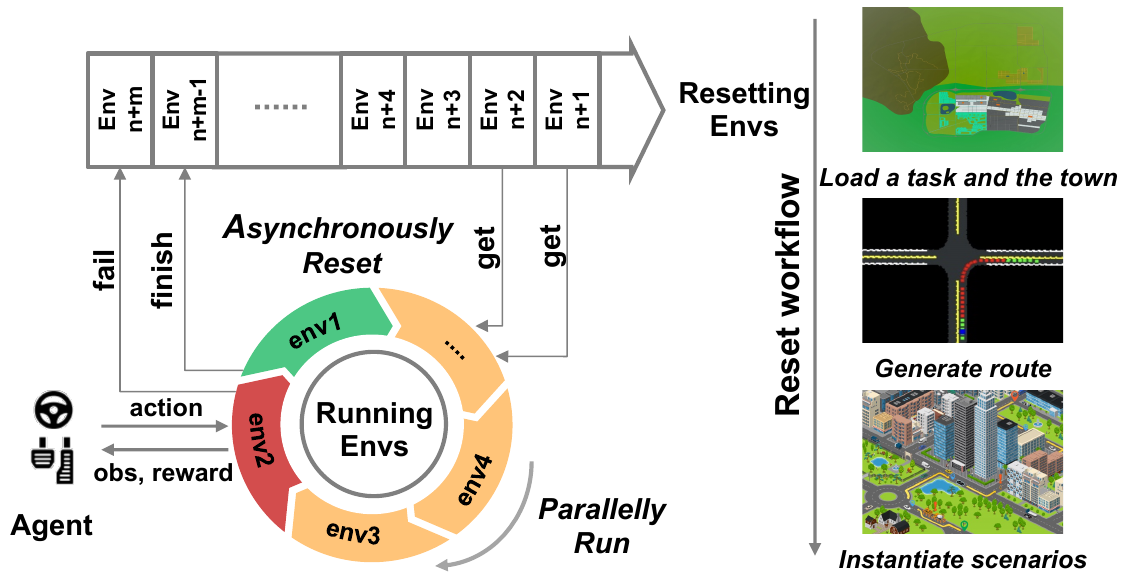}
    \caption{\textbf{Procedure of the Wrapped RL Environment}. 
    The right shows the workflow of each reset. 
    In large maps like town12 and town13, the entire reset workflow takes about 1 minute.}
    \label{fig:benchmark}

\end{figure}

We boost CARLA running efficiency via asynchronous reloading and parallel execution which could avoid the waiting time caused by the long preparation time of starting a new route, as shown in \cref{fig:benchmark}. For each 100K steps of execution, it can reduce about 1 day time cost. 

\section{Hyper-parameters}
\begin{table}[h]
  \centering
  \caption{\textbf{Hyper-parameters of the neural network}. The CNN encoder maps the $128 \times 128$ BEV to the $4 \times 4$ feature map by $4 \times4$ convolutional kernel with stride 2. The flattened feature maps, concatenated with the state features output by the MLP encoder, are then input to the RSSM which consists of GRU cells and a few dense layers. The structure of the decoder inverts that of the encoder and outputs the reconstructed mask and state vector. Please refer to DreamerV3~\cite{hafner2023mastering} for details about network.}
  \begin{tabular}{cccccc}
    \toprule
    GRU recurrent units & CNN multiplier & Dense hidden units & MLP layers & Parameters \\
    \midrule
    512 & 96 & 512 & 5 & 104M \\
    \bottomrule
  \end{tabular}
  \label{tab:model-size}
\end{table}

\end{document}